\documentclass[twoside]{article}

\usepackage[preprint]{aistats2027}
\usepackage[round]{natbib}

\usepackage[T1]{fontenc}
\usepackage[utf8]{inputenc}
\usepackage{graphicx}
\usepackage{amsmath}
\usepackage{amsfonts}
\usepackage{amsthm}
\usepackage{amssymb}
\usepackage{booktabs}
\usepackage{multirow}
\usepackage{xcolor}
\usepackage{tikz}
\usetikzlibrary{arrows.meta,positioning}
\usepackage{hyperref}
\usepackage{url}

\usepackage{amsmath,amsfonts,bm}

\def\eqref#1{equation~\ref{#1}}

\def\1{\bm{1}}

\DeclareMathAlphabet{\mathsfit}{\encodingdefault}{\sfdefault}{m}{sl}
\SetMathAlphabet{\mathsfit}{bold}{\encodingdefault}{\sfdefault}{bx}{n}

\newtheorem{theorem}{Theorem}[section]

\newtheorem{lemma}[theorem]{Lemma}
\newtheorem{proposition}[theorem]{Proposition}

\definecolor{dark2green}{rgb}{0.1, 0.65, 0.3}
\definecolor{dark2orange}{rgb}{0.9, 0.4, 0.}
\definecolor{dark2purple}{rgb}{0.4, 0.4, 0.8}
\newcommand{\first}[1]{\textbf{\textcolor{dark2green}{#1}}}
\newcommand{\second}[1]{\textbf{\textcolor{dark2orange}{#1}}}
\newcommand{\third}[1]{\textbf{\textcolor{dark2purple}{#1}}}

\begin{document}

%

%

\twocolumn[

\aistatstitle{NearFormer: Linear-Size Sparse Attention via Randomized Near Neighbors}


\aistatsauthor{ Ali Parviz$^{\ast}$ \And Alex Cloninger$^{\ast,\dagger}$ \And Gal Mishne$^{\ast}$ }

\aistatsaddress{ $^{\ast}$Halicio\u{g}lu Data Science Institute, UC San Diego \\
                 $^{\dagger}$Department of Mathematics, UC San Diego } ]

\begin{abstract}
Graph transformers can capture long-range interactions, but their quadratic attention cost limits their use on large graphs. We introduce \emph{NearFormer}, a graph transformer whose global attention is restricted to a randomized near-neighbor (NN) graph. NearFormer uses an efficient randomized construction to connect each node to a small number of near neighbors from a broader neighborhood defined by a label-independent feature or spectral geometry. 
For a fixed retained budget \(k\), the resulting attention support has linear size \(\mathcal{O}(|E|+nk)\). 
The NN graph is built once by within-bin search and then held fixed. We use these edges as sparse global-attention supports, with optional learned attention biases, while preserving the original graph for local message passing. Experiments on long-range graph benchmarks and synthetic structural tasks show that NearFormer provides a favorable accuracy--scalability trade-off relative to local message passing, dense and sparse attention.

\end{abstract}

\section{Introduction}
\label{sec:introduction}
Graph learning supports applications involving relational data,
including molecular systems, social networks, and knowledge graphs. Most graph
neural networks (GNNs) follow the message-passing paradigm, in which a node
iteratively aggregates information from its local neighborhood
\citep{gilmer2017neural,hamilton2017inductive}. Although this locality provides
a useful inductive bias, information between distant nodes must traverse many
layers. Deep message passing can suffer from over-smoothing
\citep{OonoSuzuki20} and over-squashing
\citep{alon2020bottleneck,topping2022riccurvature}. Its distinguishing power is also generally
bounded by the 1-Weisfeiler--Lehman test
\citep{morris2019weisfeiler,xu2018powerful}.

Graph transformers offer a complementary approach by allowing attention to
model long-range interactions directly. Recent architectures encode graph
structure through positional or structural representations and combine global
attention with local message passing
\citep{kreuzer2021rethinking,Ying2021DoTR,MialonCSM21,RampasekGDLWB22,black2024comparing}.
Dense self-attention provides this flexibility by forming all node pairs,
which requires $\mathcal{O}(n^2)$ time and memory
for a graph with $n$ nodes. This cost is manageable for small molecular graphs
but becomes prohibitive on larger graphs. Generic sparse or linear-attention
methods such as BigBird and Performer reduce this asymptotic burden
\citep{ZaheerGDAAOPRWY20,ChoromanskiLDSG21}, yet their sparsity patterns were
developed primarily for sequences and need not reflect the geometry of an
arbitrary graph.

Graph-specific sparse attention should add long-range communication channels
with few edges, and those channels should reflect graph data rather than an
arbitrary node ordering. Existing methods reduce the edge count with
data-independent random expander overlays \citep{shirzad2023exphormer} or
modify the input topology according to
curvature, spectral, or effective-resistance criteria
\citep{karhadkar2023fosr,banerjee2022oversquashing,black2023understanding,
arnaiz2022diffwire}. Expander overlays are efficient but do not adapt to the
geometry of a particular graph, whereas graph rewiring can require costly
structural optimization and changes the graph on which local message passing
is defined.

Here, we introduce \emph{NearFormer}, a scalable graph transformer based on
\emph{randomized near-neighbor} (NN) attention. We construct a near-neighbor graph in a task-relevant, label-independent geometry, where each node retains $k$ randomly selected neighbors from a broader neighborhood at the scale of its $K$ nearest neighbors \citep{linderman2020randomized}. 
The geometry may be defined by node attributes or graph Laplacian
eigenvectors.
Retaining $k$ neighbors per node yields at most $2nk$ directed edge instances
after symmetrization, and hence linear edge complexity for fixed $k$. 
In
contrast to $K$-\emph{nearest-neighbor} graphs, NN separates the size
of the conceptual neighborhood from the number of retained edges: it obtains
a randomized sample at that geometric scale without constructing all $K$
nearest-neighbor connections.

NearFormer preserves the original edge set for local message passing and uses
NN edges as a separate relation for long-range attention. This separation
avoids conflating a learned communication pattern with the physical or
semantic topology of the input graph. The NN edges can be used as a hard
sparse-attention support with an optional bounded, learnable logit bias, allowing
the model to suppress or favor NN connections without densifying attention.

\noindent\textbf{Main contributions.}
We propose randomized near-neighbor attention, a label-independent
construction that samples $k$ edges at each node from a geometric neighborhood of scale $K$, yielding linear edge complexity for fixed $k$.  NearFormer
constructs this graph using feature or spectral geometry and
employs NN edges as sparse global-attention supports or learned
biases while preserving the input graph for local message passing.  We
evaluate NearFormer on long-range graph benchmarks and synthetic
structural tasks against dense and sparse attention and local message passing, characterizing its accuracy-scalability trade-off.

\section{Background}
\label{sec:background}
\noindent\textbf{Graphs and message passing.}
Let $G=(V,E)$ be an undirected graph with $n=|V|$ nodes, adjacency matrix $\mathbf{A}\in\mathbb{R}^{n\times n}$, and degree matrix $\mathbf{D}=\operatorname{diag}(d_1,\ldots,d_n)$. Its symmetric normalized
Laplacian is $\mathbf{L}=\mathbf{I}-\mathbf{D}^{-1/2}\mathbf{A}\mathbf{D}^{-1/2}$. A message-passing neural network (MPNN) updates each node from its graph neighborhood $\mathcal{N}_G(v)$. After $L$ layers, information travels at most $L$ hops. When many distant signals must pass through a narrow cut, they are compressed into fixed-width representations, a phenomenon known as \emph{over-squashing} \citep{alon2020bottleneck,topping2022riccurvature,black2023understanding}.

\noindent\textbf{Attention graphs.}
A graph attention layer is specified by an interaction graph $H=(V,E_H)$: node $v$ attends only to neighbors in $H$. Dense graph transformers take $E_H=V\times V$ and require $\mathcal{O}(n^2)$ pairwise scores. Sparse transformers instead choose $|E_H|\ll n^2$, ideally $\mathcal{O}(n)$, so the interaction graph is both a computational device and an inductive bias. A related line of work constructs sparse attention from $k$-nearest-neighbor graphs, including in general deep learning 
\citep{lee2022knn,xie2020point,meng2025mvgnn}. Hybrid architectures such as GraphGPS retain $E$ for a local MPNN branch and use a separate graph for global attention \citep{RampasekGDLWB22}. NearFormer follows this distinction: the original topology governs local message passing, while a randomized near-neighbor graph supplies sparse, geometry-aware attention connections. Extended definitions and a detailed discussion of related architectures appear in Appendix~\ref{sec:related-work}.

\section{Near-Neighbor Attention}
\label{sec:nnn-gnn}
\label{sec:NearFormer-architecture}

\begin{figure*}[t]
    \centering
    \centering
        \begin{tikzpicture}[
        font=\scriptsize,
        >=Latex,
        gnode/.style={
            circle, draw=black!70, fill=white,
            minimum size=3.2mm, inner sep=0pt
        },
        gnodehl/.style={
            circle, draw=dark2purple!85!black, fill=dark2purple!18,
            minimum size=3.6mm, inner sep=0pt
        },
        block/.style={
            draw=black!60, rounded corners=1.5pt, align=center,
            minimum height=6mm, inner sep=1.8pt, fill=gray!7
        },
        add/.style={
            circle, draw=black!75, fill=gray!12,
            minimum size=5.2mm, inner sep=0pt
        },
        flow/.style={-{Latex[length=1.4mm,width=1.0mm]},semithick,    draw=black!75},
        local/.style={semithick, draw=blue!65!black},
        localv/.style={very thick, draw=blue!65!black},
        candidate/.style={densely dotted, draw=black!38},
        nnedge/.style={very thick, draw=dark2orange},
        virtedge/.style={densely dashed, draw=dark2purple!75},
        repbox/.style={
            draw=black!35, rounded corners=2pt, fill=gray!3
        },
        manifold/.style={
            draw=gray!45, densely dotted, fill=gray!10
        },
        candregion/.style={
            draw=gray!55, densely dotted, fill=gray!18, fill opacity=0.35
        }
]

\path[use as bounding box] (-1.35,-1.35) rectangle (16.45,5.05);

\node[font=\footnotesize\bfseries] at (4.0,4.85)
{(a) Randomized near-neighbor support construction};

\begin{scope}[shift={(1.5,3.1)},scale=0.95]
    \node[gnodehl] (g1) at (-0.85,0.35) {};
    \node[gnode] (g2) at (-0.15,0.85) {};
    \node[gnode] (g3) at ( 0.75,0.45) {};
    \node[gnode] (g4) at ( 0.70,-0.45) {};
    \node[gnode] (g5) at (-0.20,-0.75) {};
    \node[gnode] (g6) at (-0.95,-0.35) {};

    \draw[local] (g2)--(g3)--(g4)--(g5)--(g6);
    \draw[local] (g2)--(g6);
    \draw[localv] (g6)--(g1)--(g2);
\end{scope}
\node at (1.5,2.00) {input $G=(V,E)$};

\begin{scope}[shift={(1.5,0.35)},scale=1.0]

    \filldraw[manifold]
        (-1.95,-0.58)
        .. controls (-1.55,0.55) and (-0.75,0.90) .. (0.18,0.74)
        .. controls (0.98,0.60) and (1.45,0.12) .. (1.58,-0.34)
        .. controls (1.28,-0.88) and (0.38,-0.98) .. (-0.46,-0.84)
        .. controls (-1.08,-0.75) and (-1.58,-0.72) .. cycle;

    \node[gnodehl] (rv)  at (-1.18,-0.08) {};
    \node[gnode]   (r1)  at (-1.35, 0.42) {};
    \node[gnode]   (r2)  at (-0.83,-0.48) {};
    \node[gnode]   (r3)  at (-0.62, 0.58) {};
    \node[gnode]   (r4)  at (-0.28, 0.10) {};
    \node[gnode]   (r5)  at ( 0.02,-0.56) {};
    \node[gnode]   (r6)  at ( 0.18, 0.50) {};
    \node[gnode]   (r7)  at ( 0.52,-0.08) {};
    \node[gnode]   (r8)  at ( 0.82, 0.42) {};
    \node[gnode]   (r9)  at ( 1.05,-0.38) {};
    \node[gnode]   (r10) at ( 1.26, 0.02) {};

    \draw[candidate] (rv)--(r1);
    \draw[candidate] (rv)--(r2);
    \draw[candidate] (rv)--(r3);
    \draw[candidate] (rv)--(r4);
    \draw[candidate] (rv)--(r5);
    \draw[candidate] (rv)--(r6);
    \draw[candidate] (rv)--(r7);
    \draw[candidate] (rv)--(r8);
    \draw[candidate] (rv)--(r9);
    \draw[candidate] (rv)--(r10);

    \node[gnode, fill=dark2orange, draw=dark2orange] at (r6) {};
    \node[gnode, fill=dark2orange, draw=dark2orange] at (r9) {};

    \node[font=\tiny, anchor=east] at (rv.west) {$v$};
    \node[font=\tiny] at (0.10,0.92) {$Z$};
\end{scope}

\begin{scope}[shift={(5,0.35)},scale=0.95]
    \node[gnodehl] (n1) at (-0.85,0.35) {};
    \node[gnode]   (n2) at (-0.15,0.85) {};
    \node[gnode] (n3) at ( 0.75,0.45) {};
    \node[gnode]   (n4) at ( 0.70,-0.45) {};
    \node[gnode] (n5) at (-0.20,-0.75) {};
    \node[gnode]   (n6) at (-0.95,-0.35) {};
    \node[gnode, fill=dark2orange, draw=dark2orange] at (n5) {};
    \node[gnode, fill=dark2orange, draw=dark2orange] at (n3) {};

    \draw[candidate] (n1)--(n2)--(n3)--(n4)--(n5)--(n6)--(n1);
    \draw[candidate] (n1)--(n3) (n1)--(n4) (n1)--(n5);
    \draw[candidate] (n2)--(n4) (n2)--(n5) (n2)--(n6);
    \draw[candidate] (n3)--(n5) (n3)--(n6) (n4)--(n6);

    \draw[nnedge] (n1)--(n3);
    \draw[nnedge] (n1)--(n5);

    \node[font=\tiny, anchor=east] at (n1.west) {$v$};

\end{scope}
\node[align=center] at (5.2,-0.65)
{$E_{\mathrm{NN}} $};
  \coordinate (EHout) at (7.9,1.95);
\begin{scope}[shift={(7,1.95)},scale=0.95]
    \node[gnodehl] (h1) at (-0.85,0.35) {};
    \node[gnode]   (h2) at (-0.15,0.85) {};
    \node[gnode] (h3) at ( 0.75,0.45) {};
    \node[gnode]   (h4) at ( 0.70,-0.45) {};
    \node[gnode] (h5) at (-0.20,-0.75) {};
    \node[gnode]   (h6) at (-0.95,-0.35) {};

    \draw[local] (h2)--(h3)--(h4)--(h5)--(h6);
    \draw[local] (h2)--(h6);
    \draw[localv] (h6)--(h1)--(h2);

    \draw[nnedge] (h1)--(h3);
    \draw[nnedge] (h1)--(h5);

\end{scope}
\node[align=center] at (6.8,3.1)
{$E_H = E \cup E_{\mathrm{NN}} $};


\draw[flow] (3.15,0.35) -- (3.85,0.35);

\draw[flow] (2.45,3.00) -- (6.0,2.10);

\draw[flow] (5.15,1.1) -- (5.90,1.70);

    \begin{scope}[xshift=-3.6cm]
    \node[font=\footnotesize\bfseries] at (15.7,4.85)
    {(b) Local--global NearFormer layer $\ell$};

    \node (xin) at (15.7,-1.15) {$\mathbf{X}^{\ell}$};

    \node[block, fill=yellow!45, text width=2.4cm] (attn)
    at (13.5,-0.05) {sparse multi-head attention\\support: $E_H$};

    \node[block, fill=dark2purple!18, text width=2.4cm] (mpnn)
    at (18.35,-0.05) {local MPNN\\support: $E$};

    \draw[flow] (xin.north) to[out=110,in=270] (attn.south);
    \draw[flow] (xin.north) to[out=70,in=270] (mpnn.south);

    \node[add] (mplus) at (13.5,1.05) {$+$};
    \node[add] (aplus) at (18.35,1.05) {$+$};

    \draw[flow] (attn.north) -- node[right,font=\scriptsize] {dropout} (mplus.south);
    \draw[flow] (mpnn.north) -- node[right,font=\scriptsize] {dropout} (aplus.south);

    \draw[flow] (xin.west) -| (11.95,1.05) -- (mplus.west);
    \draw[flow] (xin.east) -| (19.75,1.05) -- (aplus.east);

    \node[anchor=west] at (13.65,1.05) {residual + norm};
    \node[anchor=west] at (15.90,1.05) {residual + norm};

    \node (xm) at (13.5,1.70) {$\mathbf{X}^{\ell+1}_{M}$};
    \node (xt) at (18.35,1.70) {$\mathbf{X}^{\ell+1}_{T}$};+
    
    \draw[flow] (mplus.north) -- (xm.south)+(0,0.08);
    \draw[flow] (aplus.north) -- (xt.south)+(0,0.08);

    \node[add] (sum) at (15.7,2.10) {$+$};
    \draw[flow] (xm.east) to[out=10,in=190] (sum.west);
    \draw[flow] (xt.west) to[out=170,in=-10] (sum.east);
    \node[anchor=west] at (16.05,2.25) {branch sum};

    \node[block, fill=blue!15, text width=2.9cm] (ffn)
    at (15.7,2.90) {two-layer feedforward network};
    \draw[flow] (sum.north) -- (ffn.south);

    \node[add] (fplus) at (15.7,3.70) {$+$};
    \draw[flow] (ffn.north) -- node[right,font=\scriptsize] {dropout} (fplus.south);

    \draw[flow] (sum.west) -| (13.35,3.70) -- (fplus.west);
    \node[anchor=west] at (16.05,3.70) {residual + norm};

    \node (xout) at (15.7,4.30) {$\mathbf{X}^{\ell+1}$};
    \draw[flow] (fplus.north) -- (xout.south)+(0,0.08);
\end{scope}

\draw[flow] (EHout) to[out=0,in=180] (attn.west);
    \end{tikzpicture}%
    \caption{Randomized near-neighbor attention in one NearFormer layer.
    \textbf{(a)}~$E_{\mathrm{NN}}$ connects each~$v$ to $k$ random near
    neighbors in representation~$Z$; $E_H=E\cup E_{\mathrm{NN}}$.
    \textbf{(b)}~Local MPNN on~$E$ and sparse attention on~$E_H$ are
    residual-normalized, summed, and passed through a residual FFN.}
    \label{fig:NearFormer-architecture}
\end{figure*}
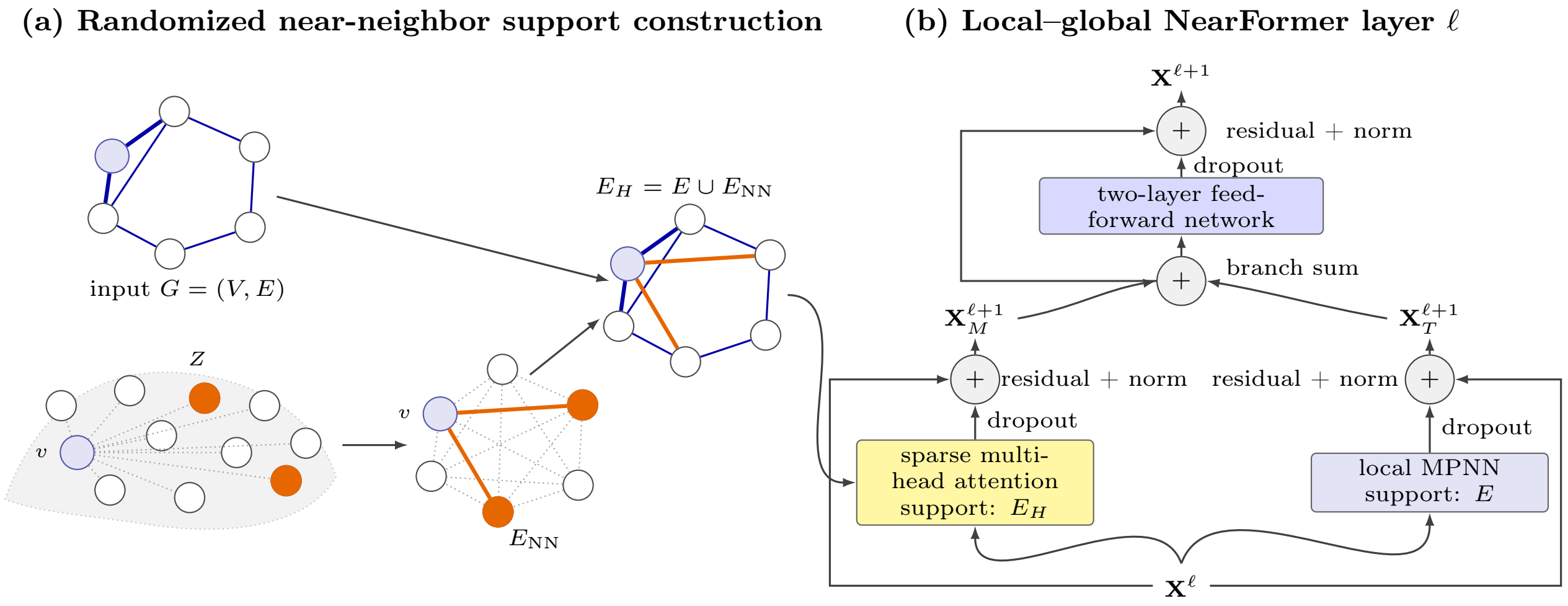

We define a graph-transformer layer using an \emph{interaction graph}
$H$, whose directed edges specify the node pairs for which attention scores
are computed. Dense self-attention evaluates $\mathcal{O}(n^2)$ scores over
all directed pairs. NearFormer instead augments the input-graph topology with
sparse, geometry-dependent interactions.
Figure~\ref{fig:NearFormer-architecture} shows how the sampled NN edges define
the sparse support used by the global attention branch.

For input graph $G=(V,E)$ and node features $Z$,
we use both orientations of
each input edge when $E$ is supplied as an undirected edge set. The NearFormer
interaction graph contains two types of connections:
\begin{enumerate}
    \item \textbf{Local neighborhood attention.}
    The edges in $E$ preserve the relational structure of the input and allow
    attention or message passing between adjacent nodes. These connections
    contribute $\mathcal{O}(|E|)$ interactions.

    \item \textbf{Randomized near-neighbor attention.}
    The edge set $E_{\mathrm{NN}}$ connects nodes that are close in a
    prespecified representation space constructed without target labels. Each node is connected to approximately $k$ random nodes out of its $K$-nearest neighborhood, with $k\ll K$.
    The resulting edges can be \emph{nonlocal} in the input graph even though they are local in the chosen representation.

\end{enumerate}
After adding reverse edges, the resulting support is
\begin{equation}
    E_H
    =
        E\cup E_{\mathrm{NN}}.
    \label{eq:NearFormer-support}
\end{equation}
Self-loops, when required by a particular attention implementation, are added
separately. For a fixed NN budget $k$, sparse
attention on this support evaluates
$\mathcal{O}(|E|+nk)=\mathcal{O}(|E|+n)$ attention interactions. This
bound concerns the attention layer. As in
Exphormer~\citep{shirzad2023exphormer}, attention is
restricted to a sparse graph. NearFormer replaces Exphormer's data-independent
random expander edges with the graph-dependent near-neighbor edges
$E_{\mathrm{NN}}$, which provide its long-range interactions.

\subsection{Randomized near-neighbor graph}
\label{sec:NN-construction}
A na\"ive NN graph construction would require computing the K-nearest neighbors of each node. We instead use an efficent  randomized binning scheme. 
Let $z_i\in\mathbb{R}^{d_z}$ denote the fixed representation used to compare
node $v_i$ with other nodes. Given a requested candidate budget $K_0$, set
$K=\min\{K_0,n-1\}$. For a retained degree $k\leq K$, set
$M=\max\{1,\lfloor K/k\rfloor\}$. We draw two independent random
permutations of $V$ and divide each into $M$ nearly equal bins. The first
permutation defines target bins $B_1,\ldots,B_M$, while the second defines
query bins $Q_1,\ldots,Q_M$. Thus every node occurs once as a target and,
independently, once as a query. For each $v_i\in Q_b$, define its retained
neighbors as the $k$ nearest targets in the paired bin $B_b$:
\begin{equation}
    \mathcal{N}_{b,k}(v_i)
    =
    \underset{\substack{S\subseteq B_b\setminus\{v_i\}\\
                        |S|=\min\{k,|B_b\setminus\{v_i\}|\}}}
    {\arg\min}
    \sum_{v_j\in S} d(z_i,z_j),
    \label{eq:NN-bin-neighbors}
\end{equation}
where $d$ is either squared Euclidean distance or cosine distance and ties
may be broken arbitrarily. We symmetrize these pairs to obtain $E_{\mathrm{NN}}$. 
Each node contributes at most $k$ unsymmetrized edges, so
$E_{\mathrm{NN}}$ contains at most $2nk$ directed edges before
deduplication.

This construction adapts the subsampled-search strategy of
\citet[Algorithm~1]{linderman2020randomized}. Each target bin contains
approximately $n/M\approx nk/K$ nodes. Any member of the exact $K$-nearest
neighbors of a query lands in its assigned target bin with probability
approximately $1/M\approx k/K$; hence approximately $k$ of those $K$
neighbors are exposed to the query in expectation. Retrieving $k$-nearest
targets in that bin therefore efficiently approximates selecting a $k$-element random
subset from a neighborhood at scale $K$, without constructing a
global $K$-nearest-neighbor graph, or an
$n\times n$ distance matrix. The parameter $K$ controls this scale through
the number of bins, whereas $k$ controls the retained number of near edges.


The inductive bias of $E_{\mathrm{NN}}$ is determined by the representation
$z_i$. We consider two fixed precomputed geometries. When node attributes
provide a nondegenerate comparison space, we use the feature geometry
$z_i=x_i$. If attributes are absent or constant within a graph, feature
distances cannot distinguish its nodes. Then we use $q$
lowest-frequency eigenvectors of the symmetric normalized graph Laplacian
after omitting the first trivial mode. This choice depends only on the
observed input graph and attributes, is fixed for each dataset before
training, and does not use target labels.

The NN graph is constructed during preprocessing and remains fixed throughout training, where we use exact or approximate nearest-neighbor search inside each target bin (the preprocessing
cost is analyzed in Appendix~\ref{sec:computational-complexity}). Each vertex is queried with respect to its assigned target bin, and we concatenate the resulting edges. 
Each graph is constructed independently so no interaction is introduced between different graphs in the data.
Proposition~\ref{prop:bin-rank-selection} gives the exact rank-selection probabilities of this procedure. For the remaining expectation identities, Section~\ref{sec:NN-theory} uses an exact-candidate surrogate that samples with replacement. The surrogate is not used in the experiments.

Unless specified otherwise, we set $K_0=100$ and 
\begin{equation}
    k=\min\!\left\{K,\max\!\left\{2,\left\lfloor
    \log\!\bigl(\log(\max\{n,2\})+1\bigr)\right\rfloor\right\}\right\},
    \label{eq:NN-default-k}
\end{equation}
computed separately for each graph.
If $n=1$, then $K=0$ and no NN edge is added.


\subsection{Architectural variants}
\label{sec:NN-integration}

NearFormer distinguishes the input edges $E$ from the auxiliary edges
$E_{\mathrm{NN}}$. Retaining the edge types lets a single sparse attention operator process both relations and also supports separate local and global branches. The same representation enables controlled comparisons between attention and message passing over NN edges.

\noindent\textbf{Unified sparse attention.}
In the sparse variant, a single attention operator acts on the support
of $E_H$ (\ref{eq:NearFormer-support}). Edge-type embeddings distinguish original 
and NN connections, while dataset-provided edge features are
retained on $E$. The architecture follows a sparse graph transformer, with NN
edges replacing random expander edges. We denote the hard-support variant
by \emph{NearFormer (NN)}. We also introduce an NN-specific learned logit
bias by \emph{NearFormer (NN-bias)}, where for
head $h$ in layer $\ell$, the attention coefficient on an allowed pair may
also include an NN-specific learned bias:
\begin{equation}
    \alpha_{uv}^{(h)}
    =
    \frac{
      \exp\!\left(
        s_{uv}^{(h)}+
        \beta_\ell\,\mathbf{1}[(u,v)\in E_{\mathrm{NN}}]
      \right)}
    {\sum_{w:(w,v)\in E_H}
      \exp\!\left(
        s_{wv}^{(h)}+
        \beta_\ell\,\mathbf{1}[(w,v)\in E_{\mathrm{NN}}]
      \right)},
    \label{eq:NN-soft-bias}
\end{equation}
where $s_{uv}^{(h)}$ is the edge-conditioned query--key score and
$\beta_\ell=b_{\max}\tanh(\theta_\ell)$ is a bounded learned scalar. The
bound prevents an unconstrained additive logit from dominating the softmax.
Setting $\theta_\ell=0$ reduces to the hard-support variant. NN edges
remain in the support and retain their edge-type features. Learning
$\beta_\ell$ lets the layer favor or suppress the NN relation without
changing the sparsity pattern, so the interaction count remains linear in
$|E|+n$ for fixed $k$.

\noindent\textbf{Local--global NearFormer layers.}
NearFormer can also be instantiated in the GraphGPS framework
\citep{RampasekGDLWB22}. The local branch propagates exclusively on $E$, while
the global branch restricts attention to
$E\cup E_{\mathrm{NN}}\cup\{(v,v):v\in V\}$. Both NearFormer (NN) and
NearFormer (NN-bias) use this same sparse support; they differ only in whether
the bounded NN-specific bias in~\eqref{eq:NN-soft-bias} is applied. Thus both
variants evaluate $\mathcal{O}(|E|+nk)$ pairwise scores. With hidden states
$H^{(\ell)}$, a layer has the form
\begin{equation}
  \begin{aligned}
    H^{(\ell+1)}
    = \operatorname{FFN}_{\ell}\Bigl(
      &\operatorname{MPNN}_{\ell}(H^{(\ell)},E)\\
      &+\operatorname{Attn}_{\ell}
      (H^{(\ell)};E,E_{\mathrm{NN}})
    \Bigr),
  \end{aligned}
  \label{eq:NN-gps-layer}
\end{equation}
with residual connections and normalization in both branches.

\noindent\textbf{Near-neighbor message-passing.}
We denote by \emph{NearFormer (NN-MP)} a routing ablation that places the
sampled NN relation in the local branch rather than the attention branch.
Specifically, NearFormer (NN) and NearFormer (NN-bias) keep the local MPNN
support fixed at $E$ and include $E_{\mathrm{NN}}$ in sparse attention,
whereas NearFormer (NN-MP) applies the MPNN to
$E\cup E_{\mathrm{NN}}$ and excludes $E_{\mathrm{NN}}$ from attention.
Thus NearFormer (NN-MP) is distinct from both preceding variants; it changes
where NN messages are computed rather than changing their attention logits.
NN edges receive a separate learned edge-type embedding. We evaluate this
variant separately because merging the relations changes the graph used for
local aggregation and may overwhelm a task-relevant physical graph. The
comparison distinguishes gains from sparse long-range attention from those
caused by modifying local message passing.

\section{Theoretical Analysis}
\label{sec:theoretical-analysis}
\subsection{NN attention as geometry-aware rewiring}
\label{sec:NN-theory}

Over-squashing and over-smoothing impose competing requirements on an
interaction graph.  Over-squashing arises when information from many distant
nodes must traverse long paths or narrow cuts before being compressed into
fixed-width states \citep{alon2020bottleneck,topping2022riccurvature,
black2023understanding}.  Adding edges can shorten these routes and reduce
topological bottlenecks.  By contrast, over-smoothing is associated with
repeated mixing, which progressively removes variation between node
representations \citep{OonoSuzuki20,rusch2023survey}.  Indiscriminately adding
edges may therefore improve communication while accelerating global mixing.

NearFormer offers a geometry-dependent mechanism for balancing these effects.
Its edges are local in the
representation metric \(d(z_i,z_j)\), but need not be local in the input-graph
metric \(d_G(v_i,v_j)\).  Thus, an NN edge can be a long-range shortcut in \(G\)
while joining nodes that remain close in a task-relevant geometry.  The
results concern the undirected support
obtained after symmetrizing the construction in
Section~\ref{sec:NN-construction}. They neither assume uniform learned
attention weights nor imply improved prediction.

\subsection{Path compression and over-squashing}

Assume that \(G=(V,E)\) is a connected, simple, undirected graph.  Write
\(E_{\mathrm{NN}}\) for the undirected support of the directed relation
constructed in Section~\ref{sec:NN-construction}: the set of distinct pairs
\(\{v,u\}\) for which at
least one of \((v,u)\) or \((u,v)\) is sampled.  Let
\(H=(V,E\cup E_{\mathrm{NN}})\) and assign unit conductance
to every edge of \(H\).  Write \(d_H\) and \(R_H\) for shortest-path and
effective-resistance distance in \(H\), respectively.  Throughout,
the candidate-set size \(K\) and the per-node budget \(k\) are those of
Section~\ref{sec:NN-construction}.

\begin{lemma}[Monotonicity under NN augmentation]
\label{lem:NN-monotonicity}
For every \(x,y\in V\),
\[
    d_H(x,y)\leq d_G(x,y),
    \qquad
    R_H(x,y)\leq R_G(x,y).
\]
Moreover, if \(\{v,u\}\in E_{\mathrm{NN}}\), then
\[
 d_H(x,y)\leq
 d_G(x,v)+1+d_G(u,y).
\]
\end{lemma}
The proof is given in Appendix~\ref{sec:NN-lemma-proofs}.
Lemma~\ref{lem:NN-monotonicity} relates NN augmentation to two topological
accounts of over-squashing: long propagation distances
\citep{alon2020bottleneck,di2023over} and large effective resistance
\citep{black2023understanding}. It characterizes communication capacity but
does not guarantee bounds on the Jacobian or prediction error of a trained
network. It applies directly to the NN message-passing support. For attention,
it applies only to the support because a learned weight may be negligible.

\subsection{Selection probabilities of the bin construction}
\label{sec:NN-bin-selection}

The bin construction of Section~\ref{sec:NN-construction} does not sample
uniformly from a fixed candidate list. It nevertheless induces a
rank-dependent selection distribution that can be computed exactly. Fix a
query \(v_i\) and order \(V\setminus\{v_i\}\) by increasing distance
\(d(z_i,\cdot)\), breaking ties by a fixed rule that does not depend on the
random permutations. Write \(v_{i,(r)}\) for the node of rank
\(r\in\{1,\ldots,n-1\}\), and let \(s_1,\ldots,s_M\) be the bin sizes, which
are shared by the target and query partitions.

\begin{proposition}[Rank selection under binning]
\label{prop:bin-rank-selection}
Before symmetrization, the probability that \(v_{i,(r)}\) is a retained
outgoing neighbor of \(v_i\) is
\[
  \psi_{n,k}(r)
  =
  \sum_{b=1}^{M}\frac{s_b}{n}
  \sum_{j=0}^{k-1}
  \frac{\binom{r-1}{j}\binom{n-r}{s_b-1-j}}{\binom{n}{s_b}},
\]
where \(\binom{a}{c}=0\) whenever \(c<0\) or \(c>a\). Moreover,
\(\sum_{r=1}^{n-1}\psi_{n,k}(r)
=\mathbb{E}\bigl[\min\{k,|B_b\setminus\{v_i\}|\}\bigr]\leq k\), where
\(B_b\) is the target bin paired with the query bin of \(v_i\).
\end{proposition}

The proof is given in Appendix~\ref{sec:NN-lemma-proofs}. The geometry
enters only through the rank \(r\): a node is retained exactly when it falls
in the assigned target bin together with fewer than \(k\) closer nodes. For
\(r\leq k\), \(\psi_{n,k}(r)=\sum_b(s_b/n)^2\), which equals \(1/M\) when all
bins have size \(n/M\). For equal bins and \(s_b\ll n\), approximating the
hypergeometric count by a binomial with \(p=1/M\):
\[
  \psi_{n,k}(r)
  \approx
  p\,\Pr\bigl[\operatorname{Bin}(r-1,p)\leq k-1\bigr].
\]
The selection probability therefore stays near \(1/M\approx k/K\) while the
expected number \((r-1)/M\) of closer nodes in the bin is below \(k\), and it
decays rapidly beyond the rank scale \(kM\), where
\(K-k<kM\leq K\) because \(M=\lfloor K/k\rfloor\). Binning thus acts as a soft
rank cutoff at scale \(K\) while retaining about \(k\) neighbors per query:
\(K\) sets how far in representation the construction reaches, and \(k\)
sets how many edges it keeps.

The same expression describes the small-graph limit. When \(K=n-1\), the
rank scale \(kM\) covers essentially the whole graph and each bin contains
only about \(k\) or \(k+1\) nodes, so nearly every eligible member of the
assigned bin is retained. The profile \(\psi_{n,k}\) is then close to flat,
and the NN relation behaves like a sparse global overlay with only weak
geometric bias. Results on graphs with \(n\lesssim K_0\) should be
interpreted accordingly.

\subsection{An exact-candidate surrogate}

To obtain closed-form expressions for shortcut gain and candidate impurity,
the remaining results analyze a simpler surrogate in which each node draws
uniformly from its exact \(K\) nearest neighbors. In this surrogate, every
rank \(r\leq K\) is drawn with probability \(1-(1-1/K)^k\approx k/K\) and no
rank beyond \(K\) is drawn. It replaces the soft profile of
Proposition~\ref{prop:bin-rank-selection} by a hard cutoff at the same scale
and per-rank level. The identities below are exact for the surrogate and
approximate for the bin construction. Let
\begin{equation}
    \mathcal{C}_K(v_i)
    =
    \underset{\substack{S\subseteq V\setminus\{v_i\}\\ |S|=K}}
    {\arg\min}
    \sum_{v_j\in S} d(z_i,z_j),
    \label{eq:NN-candidates}
\end{equation}
and write
\(\mathcal{C}_K(v_i)=\{v_{\nu_i(1)},\ldots,v_{\nu_i(K)}\}\), where
\(\nu_i(j)\) is the global node index of the \(j\)th candidate. For each
node \(v_i\), let \(J_{i,1},\ldots,J_{i,k}\) be independent uniform draws
from \(\{1,\ldots,K\}\), independently across nodes. The resulting
with-replacement edge set is
\begin{equation}
    \widetilde E_{\mathrm{NN}}
    =
    \bigcup_{v_i\in V}\bigcup_{r=1}^{k}
    \left\{
      (v_i,v_{\nu_i(J_{i,r})}),
      (v_{\nu_i(J_{i,r})},v_i)
    \right\}.
    \label{eq:NN-ideal-edges}
\end{equation}
The identities below apply to \(\widetilde E_{\mathrm{NN}}\), not exactly
to the bin-based implementation used in the experiments.

The motivating theorem of \citet{linderman2020randomized} instead retains
each of \(K=c_{d,1}\log n\) exact nearest-neighbor edges independently with
probability \(p=c_{d,2}\log\log n/\log n\). Under its i.i.d.\ uniform point
model, the resulting graph has a component of size \(n-o(n)\) with high
probability and expected edge count \(\Theta(n\log\log n)\). This result does
not imply expansion, reduced over-squashing, or improved prediction for our
finite graph-derived representations.

The idealized construction admits an exact edge-local measure of its
expected path compression. Specifically,
\(E_{\mathrm{NN}}\) denotes the undirected support of
\(\widetilde E_{\mathrm{NN}}\) in~\eqref{eq:NN-ideal-edges}, and
\(H=(V,E\cup E_{\mathrm{NN}})\). For an unordered pair
\[
 m_{ij}
 =\mathbf{1}[v_j\in\mathcal{C}_K(v_i)]
  +\mathbf{1}[v_i\in\mathcal{C}_K(v_j)] ,
\]
where \(\mathbf{1}[\cdot]\) is the indicator function.  Thus
\(m_{ij}\in\{0,1,2\}\) records the number of candidate-set directions for
the pair.  Let
\(\mathcal{U}_K=\{\{v_i,v_j\}\in\binom{V}{2}:m_{ij}>0\}\).  The pair enters
\(E_{\mathrm{NN}}\) if and only if at least one of the \(km_{ij}\) eligible
draws selects the opposite endpoint, so
\[
 \pi_{ij}
 :=\Pr\bigl(\{v_i,v_j\}\in E_{\mathrm{NN}}\bigr)
 =1-\left(1-\frac{1}{K}\right)^{km_{ij}}.
\]
The realized edge-local compression is
\[
 S(H;G)
 =\sum_{\{v_i,v_j\}\in\mathcal{U}_K}
   \bigl(d_G(v_i,v_j)-1\bigr)
   \mathbf{1}[\{v_i,v_j\}\in E_{\mathrm{NN}}].
\]
Pairs already present in \(E\) contribute zero, since then
\(d_G(v_i,v_j)=1\).

\begin{lemma}[Expected edge-local compression]
\label{lem:NN-compression}
Under the exact with-replacement construction,
\[
 \mathbb{E}[S(H;G)]
  =\sum_{\{v_i,v_j\}\in\mathcal{U}_K}
   \bigl(d_G(v_i,v_j)-1\bigr)\pi_{ij}.
\]
Moreover,
\[
\sum_{\{v_i,v_j\}\in\binom{V}{2}}
 \bigl(d_G(v_i,v_j)-d_H(v_i,v_j)\bigr)
 \geq S(H;G).
\]
\end{lemma}

The proof is given in Appendix~\ref{sec:NN-lemma-proofs}.
This result shows the effect of choosing \(K\) independently of \(k\). A broad
candidate set can contain pairs that are close in representation space but
far apart in \(G\), producing large edge-local compression, while \(k\)
controls the retained degree and attention cost. Shortcut gain must be weighed against mixing across task-distinct regions.
Appendix~\ref{sec:NN-purity} defines a candidate impurity \(q_i\) and gives
the expected number of cross-region draws under the surrogate
(Lemma~\ref{lem:NN-purity}). Appendix~\ref{sec:NN-region-cuts} then relates
those draws to conductance and effective resistance. Those results are exact
for the surrogate and only approximate for the bin construction.

\section{Experiments}
\label{sec:experiments}
We evaluate the three architectural variants of NearFormer defined in
Section~\ref{sec:NN-integration} on three long-range benchmarks~\citep{mathys2026lrim,lrgb-2022,miglior2026can}.
\emph{NearFormer (NN)} restricts global attention to the input and sampled NN edges. \emph{NearFormer (NN-bias)} adds the bounded, learned NN-specific logit bias in~(\ref{eq:NN-soft-bias}) on the same sparse support. \emph{NearFormer (NN-MP)} instead routes the sampled NN edges through the message-passing branch and excludes them from global attention. We compare our model to baselines including local MPNNs, dense attention, and the data-independent random-$d$ overlay of Exphormer~\citep{shirzad2023exphormer}. Appendix~\ref{sec:baseline-models} describes the baseline architectures.
Appendix~\ref{sec:dataset-descriptions} reports the dataset statistics, and Appendix~\ref{sec:computational-complexity} analyzes the computational cost and scalability of NearFormer. Appendix~\ref{sec:exp-hyperparams} records the
training settings. 

\subsection{Long-Range Ising Model benchmark}
\label{sec:lrim-results}
LRIM is a node-regression task whose target provably depends on distant spins, with the interaction range set by the Ising model~\citep{mathys2026lrim}. 
LRIM represents spin configurations as periodic two-dimensional grid graphs.
Each node contains a binary spin, and the task is to predict the change in
energy produced by flipping that spin. The power-law interaction exponent
\(\sigma\) determines how strongly distant spins contribute to the target.
The hard setting uses \(\sigma=0.6\), while the easy uses
\(\sigma=1.5\). The benchmark varies the lattice width from 16 to 256, which
increases the graph size from 256 to 65,536 nodes.


We report the hard setting (\(\sigma=0.6\)) in Table~\ref{tab:lrim-baselines}, where distant spins contribute most strongly to the target. The corresponding easy results (\(\sigma=1.5\)) appear in Appendix~\ref{sec:additional-experiments} and follow the same size trend. GPS variants are strongest on the \(16\times16\) lattice, where dense attention is cheap and the required range is limited. From width 32 onward, NearFormer (NN) leads and remains practical through \(n=65{,}536\). Dense GPS is out of memory from width 64 onward. The learned-bias variant is competitive but not uniformly better, so a fixed sparse support is already effective when the interaction range is long.

\begin{table*}[t]
\centering
\begin{minipage}[t]{0.49\textwidth}
  \centering
  \caption{LRIM-hard $\mathrm{LogMSE}$~$\downarrow$ ($\sigma=0.6$).
  Easy results are in Appendix~\ref{sec:additional-experiments}.}
  \label{tab:lrim-baselines}
  \scriptsize
  \setlength{\tabcolsep}{2pt}
  \resizebox{\linewidth}{!}{%
  \begin{tabular}{@{}lccccc@{}}
  \toprule
  Model & 16 & 32 & 64 & 128 & 256 \\
  \midrule
  GIN
    & $-2.533{\pm}0.313$
    & $-2.249{\pm}0.135$
    & $-3.744{\pm}0.253$
    & $-3.672{\pm}0.171$
    & $-2.326{\pm}0.051$ \\
  GatedGCN
    & $-3.844{\pm}0.055$
    & $-4.087{\pm}0.238$
    & $-3.564{\pm}0.111$
    & \third{$-4.064{\pm}0.112$}
    & \third{$-4.285{\pm}0.126$} \\
  GatedGCN-VNG
    & \third{$-4.068{\pm}0.131$}
    & $-3.243{\pm}0.177$
    & \third{$-4.324{\pm}0.127$}
    & $-4.022{\pm}0.013$
    & $-4.076{\pm}0.006$ \\
  GPS-Base
    & \second{$-4.211{\pm}0.155$}
    & $-4.044{\pm}0.122$
    & OOM
    & OOM
    & OOM \\
  GPS-RWSE
    & $-4.011{\pm}0.129$
    & \third{$-4.134{\pm}0.075$}
    & OOM
    & OOM
    & OOM \\
  GPS-LapPE
    & \first{$-4.334{\pm}0.065$}
    & $-4.032{\pm}0.092$
    & OOM
    & OOM
    & OOM \\
  \midrule
  NearFormer (NN)
    & $-4.003{\pm}0.005$
    & \first{$-5.241{\pm}0.036$}
    & \first{$-4.925{\pm}0.025$}
    & \first{$-5.204{\pm}0.106$}
    & \first{$-5.803{\pm}0.046$} \\
  NearFormer (NN-bias)
    & $-4.027{\pm}0.013$
    & \second{$-4.432{\pm}0.092$}
    & \second{$-4.493{\pm}0.067$}
    & \second{$-5.052{\pm}0.162$}
    & \second{$-5.021{\pm}0.008$} \\
  \bottomrule
  \end{tabular}%
  }
\end{minipage}
\hfill
\begin{minipage}[t]{0.49\textwidth}
\centering
\caption{Fixed \(k=2\), varying \(K\).
LRIM-32-hard $\mathrm{LogMSE}$~$\downarrow$;
PascalVOC-SP and COCO-SP F1~$\uparrow$.}
\label{tab:knn-vs-nn-main}
\scriptsize
\setlength{\tabcolsep}{3pt}
\resizebox{\linewidth}{!}{%
\begin{tabular}{@{}lccc@{}}
\toprule
Overlay
  & LRIM-32 \(\downarrow\)
  & PascalVOC \(\uparrow\)
  & COCO \(\uparrow\) \\
\midrule
Ordinary \(k\)-NN
  & $-4.248{\pm}0.071$
  & $0.3861{\pm}0.0044$
  & $0.3388{\pm}0.0018$ \\
NearFormer, \(K=25\)
  & $-4.982{\pm}0.047$
  & $0.3965{\pm}0.0037$
  & $0.3454{\pm}0.0014$ \\
NearFormer, \(K=100\)
  & \first{$-5.241{\pm}0.036$}
  & \first{$0.4012{\pm}0.0032$}
  & \first{$0.3490{\pm}0.0012$} \\
Exphormer
  & $-4.173{\pm}0.082$
  & $0.3794{\pm}0.0048$
  & $0.3352{\pm}0.0020$ \\
\bottomrule
\end{tabular}%
}
\end{minipage}
\end{table*}

\subsection{Candidate scale at a fixed edge budget}
\label{sec:candidate-scale}
Our main design choice is to sample \(k\) edges from a neighborhood of scale \(K\), rather than to spend that budget on the \(k\) closest nodes or on geometry-independent long-range edges. Table~\ref{tab:knn-vs-nn-main} tests this claim with a matched NearFormer (NN) backbone, matched training, and
fixed retained degree \(k=2\). Ordinary \(k\)-NN is the case \(K=k\). NearFormer uses the same bin construction at \(K=25\) and at the default \(K=100\). Exphormer is a random expander of degree \(d=k\). Because \(k\) is held fixed, every method uses the same number of auxiliary attention edges. We report test \(\mathrm{LogMSE}\) on LRIM-32-hard and test F1 on PascalVOC-SP and COCO-SP, as mean\(\pm\)std over three seeds that resample both the overlay and the training run. 

On LRIM-32-hard and both superpixel datasets, performance improves as \(K\) increases, and the default neighborhood outperforms both ordinary \(k\)-NN and Exphormer. A broader geometric candidate set is therefore useful even when no extra edges are added. The full \(K\in\{10,25,50,100\}\) grid, LRIM-64-hard, and a global-random overlay are in Appendix~\ref{sec:additional-experiments}. This comparison is informative only when \(n\gg K\); on PCQM-Contact, Peptides, and ECHO the default
\(K_0=100\) saturates, so we do not interpret those results as evidence for the same mechanism.

\subsection{Long Range Graph Benchmark (LRGB)}
\label{sec:lrgb-results}
Here we compare MPNNs and graph transformers on PascalVOC-SP and COCO-SP for superpixel node classification,
Peptides-Func and Peptides-Struct for graph-level peptide prediction, and
PCQM-Contact for link prediction~\citep{lrgb-2022}. We compare local attention, Exphormer,
and the NN variants under matched training configurations. 

\begin{table*}[t]
\caption{Comparison of NearFormer variants with Long-Range Graph
Benchmark baselines. NearFormer values are test metrics
at the best-validation epoch. Best results are colored:
\first{first}, \second{second}, \third{third}.}
\label{tab:lrgb-NearFormer}
\fontsize{8.5pt}{8.5pt}\selectfont
\setlength{\tabcolsep}{4pt}
\centering
\begin{tabular}{lccccc}
\toprule
\multirow{2}{*}{\textbf{Model}}
  & \textbf{PascalVOC-SP} & \textbf{COCO-SP}
  & \textbf{Peptides-Func} & \textbf{Peptides-Struct}
  & \textbf{PCQM-Contact} \\
  & F1 score $\uparrow$ & F1 score $\uparrow$
  & AP $\uparrow$ & MAE $\downarrow$ & MRR $\uparrow$ \\
\midrule
GCN
  & $0.1268{\pm}0.0060$ & $0.0841{\pm}0.0010$
  & $0.5930{\pm}0.0023$ & $0.3496{\pm}0.0013$
  & $0.3234{\pm}0.0006$ \\
GINE
  & $0.1265{\pm}0.0076$ & $0.1339{\pm}0.0044$
  & $0.5498{\pm}0.0079$ & $0.3547{\pm}0.0045$
  & $0.3180{\pm}0.0027$ \\
GatedGCN
  & $0.2873{\pm}0.0219$ & $0.2641{\pm}0.0045$
  & $0.5864{\pm}0.0077$ & $0.3420{\pm}0.0013$
  & $0.3218{\pm}0.0011$ \\
GatedGCN+RWSE
  & $0.2860{\pm}0.0085$ & $0.2574{\pm}0.0034$
  & $0.6069{\pm}0.0035$ & $0.3357{\pm}0.0006$
  & $0.3242{\pm}0.0008$ \\
\midrule
Transformer+LapPE
  & $0.2694{\pm}0.0098$ & $0.2618{\pm}0.0031$
  & $0.6326{\pm}0.0126$ & $0.2529{\pm}0.0016$
  & $0.3174{\pm}0.0020$ \\
SAN+LapPE
  & $0.3230{\pm}0.0039$ & $0.2592{\pm}0.0158$
  & $0.6384{\pm}0.0121$ & $0.2683{\pm}0.0043$
  & $0.3350{\pm}0.0003$ \\
SAN+RWSE
  & $0.3216{\pm}0.0027$ & $0.2434{\pm}0.0156$
  & \third{$0.6439{\pm}0.0075$} & $0.2545{\pm}0.0012$
  & $0.3341{\pm}0.0006$ \\
GraphGPS
  & $0.3748{\pm}0.0109$ & $0.3412{\pm}0.0044$
  & \first{$0.6535{\pm}0.0041$}
  & \second{$0.2500{\pm}0.0005$}
  & $0.3337{\pm}0.0006$ \\
\midrule
Exphormer
  & $0.3975{\pm}0.0037$ & $0.3455{\pm}0.0009$
  & \second{$0.6527{\pm}0.0043$}
  & \first{$0.2481{\pm}0.0007$}
  & $0.3637{\pm}0.0020$ \\
\midrule
NearFormer (NN)
  & \second{$0.4012{\pm}0.0032$}
  & \second{$0.3490{\pm}0.0012$}
  & $0.5368{\pm}0.0056$ & $0.2636{\pm}0.0028$
  & \second{$0.3682{\pm}0.0024$} \\
NearFormer (NN-MP)
  & \third{$0.3980{\pm}0.0039$}
  & \third{$0.3462{\pm}0.0011$}
  & $0.6025{\pm}0.0042$
  & \third{$0.2527{\pm}0.0019$}
  & \third{$0.3645{\pm}0.0022$} \\
NearFormer (NN-bias)
  & \first{$0.4058{\pm}0.0028$}
  & \first{$0.3524{\pm}0.0010$}
  & $0.5582{\pm}0.0051$ & $0.2622{\pm}0.0025$
  & \first{$0.3720{\pm}0.0018$} \\
\bottomrule
\end{tabular}
\end{table*}

\begin{table*}[htbp]
\centering
\caption{ECHO test MAE (lower is better). A vertical rule separates
ECHO-Synth, on the left, from ECHO-Chem, on the right. Best results are colored: \first{first}, \second{second}, \third{third}.}
\label{tab:echo-results}
\scriptsize
\setlength{\tabcolsep}{2.2pt}
\resizebox{\textwidth}{!}{%
\begin{tabular}{@{}lccccc|cc@{}}
\toprule
\multirow{2}{*}{Method}
  & \multicolumn{5}{c|}{ECHO-Synth}
  & \multicolumn{2}{c}{ECHO-Chem} \\
\cmidrule(lr){2-6}\cmidrule(lr){7-8}
  & diam & ecc & sssp & Mean SSSP & Mean ER
  & Energy & Charge \\
\midrule
A-DGN
  & \third{$1.151{\pm}0.038$}
  & $4.981{\pm}0.037$
  & $1.176{\pm}0.140$
  & $4.251{\pm}0.081$
  & $3.486{\pm}0.027$
  & $12.486{\pm}1.621$
  & $6.543{\pm}0.146$ \\
DRew
  & $1.243{\pm}0.047$
  & \third{$4.651{\pm}0.020$}
  & $1.279{\pm}0.011$
  & \second{$3.425{\pm}0.107$}
  & $3.120{\pm}0.085$
  & $11.325{\pm}2.394$
  & $9.086{\pm}0.473$ \\
GraphCON
  & $2.969{\pm}0.189$
  & $5.474{\pm}0.001$
  & $5.734{\pm}0.011$
  & $5.723{\pm}0.045$
  & $5.167{\pm}0.125$
  & $14.295{\pm}0.807$
  & $19.629{\pm}0.195$ \\
GCN
  & $3.832{\pm}0.262$
  & $5.233{\pm}0.034$
  & $2.102{\pm}0.094$
  & $5.754{\pm}0.021$
  & $5.062{\pm}0.107$
  & $28.112{\pm}1.239$
  & $8.421{\pm}0.512$ \\
GCNII
  & $2.005{\pm}0.093$
  & $5.241{\pm}0.030$
  & $2.128{\pm}0.429$
  & $5.135{\pm}0.029$
  & $5.146{\pm}0.027$
  & $13.235{\pm}2.630$
  & $8.829{\pm}0.021$ \\
GIN
  & $1.630{\pm}0.161$
  & $4.869{\pm}0.092$
  & $2.234{\pm}0.271$
  & $4.307{\pm}0.084$
  & $3.988{\pm}0.061$
  & $47.851{\pm}10.154$
  & $10.784{\pm}0.059$ \\
GPS
  & $2.160{\pm}0.098$
  & $4.758{\pm}0.021$
  & \second{$0.472{\pm}0.050$}
  & $4.604{\pm}0.052$
  & \third{$2.510{\pm}0.019$}
  & \first{$5.257{\pm}0.842$}
  & $6.182{\pm}0.219$ \\
GRIT
  & \first{$1.014{\pm}0.046$}
  & $5.091{\pm}0.158$
  & \first{$0.121{\pm}0.013$}
  & $5.242{\pm}0.013$
  & $2.840{\pm}0.092$
  & $25.508{\pm}2.507$
  & $7.134{\pm}6.090$ \\
PH-DGN
  & $1.627{\pm}0.398$
  & $5.068{\pm}0.126$
  & $1.323{\pm}0.485$
  & $5.269{\pm}0.125$
  & $4.786{\pm}0.425$
  & $16.080{\pm}1.123$
  & $7.915{\pm}0.269$ \\
SWAN
  & \second{$1.121{\pm}0.070$}
  & $4.840{\pm}0.045$
  & $0.896{\pm}0.232$
  & $4.284{\pm}0.032$
  & $4.726{\pm}0.014$
  & $12.629{\pm}1.157$
  & \third{$6.109{\pm}0.103$} \\
NearFormer (NN)
  & $1.752{\pm}0.064$
  & \first{$2.396{\pm}0.031$}
  & \third{$0.559{\pm}0.024$}
  & \first{$2.752{\pm}0.064$}
  & \first{$2.105{\pm}0.024$}
  & \third{$8.427{\pm}0.106$}
  & \second{$6.004{\pm}0.245$} \\
NearFormer (NN-bias)
  & $1.92{\pm}0.128$
  & \second{$3.260{\pm}0.047$}
  & $1.101{\pm}0.401$
  & \third{$3.651{\pm}0.022$}
  & \second{$2.486{\pm}0.207$}
  & \second{$7.354{\pm}0.612$}
  & \first{$5.860{\pm}0.185$} \\
\bottomrule
\end{tabular}%
}
\end{table*}

The LRGB results separate the effects of NN edges by task family. On both superpixel datasets and on PCQM-Contact, NearFormer (NN-bias) ranks first and NearFormer (NN) ranks second. Their consistent ordering across node classification and link prediction indicates that sparse NN attention benefits from learning how strongly the auxiliary relation should enter each layer. Both variants also outperform the Exphormer on these tasks, which supports conditioning the sparse attention pattern on the input graph. The same matched-\(k\) pattern holds on the superpixel tasks (Table~\ref{tab:knn-vs-nn-main}): broadening \(K\) improves F1 at no extra edge cost, and the default neighborhood outperforms Exphormer. The Peptide tasks show a different ordering. NearFormer (NN-MP), which routes NN edges through message passing, is the strongest NearFormer variant on both Peptides-Func and Peptides-Struct. Nevertheless, all three NearFormer variants perform worse than both GraphGPS and Exphormer on both tasks. Thus, the advantage of NN edges does not transfer uniformly across task families: attention-based integration is most effective on the superpixel and contact benchmarks, while direct propagation is preferable on Peptides only among the three NearFormer variants.

\subsection{ECHO benchmark}
\label{sec:echo-synth-results}

ECHO evaluates long-range propagation in controlled synthetic graphs and
DFT-labeled molecular data~\citep{miglior2026can}. ECHO-Synth contains 10,080 graphs from six topology families: lines, ladders, grids, trees, caterpillars, and lobsters. Their diameters range from 17 to 40. Lines, trees, caterpillars, and lobsters route communication between distant nodes through narrow paths. ECHO-Chem contains approximately 170,000 molecules for atom-level partial-charge prediction and 196,000 molecules for graph-level total-energy prediction, with diameters between 17 and 40. Both targets are computed from density functional theory. 


\noindent\textbf{ECHO-Synth results. }
This benchmark includes three tasks that distinguish source-conditioned, node-level, and graph-level distance prediction.
\emph{SSSP} uses
one marked source for shortest path distance, \emph{diameter} is the maximum pairwise distance, and \emph{eccentricity}
is the maximum node-wise distance. We added two post-hoc node-level targets which average distances from
each node to all other nodes without using a marked source or a maximum.
Let
\(d_G(u,v)\) denote shortest-path distance. \emph{Mean SSSP} is the average distance
from \(v\) to all other nodes:
\[
  \operatorname{MeanSSSP}(v)
  = \frac{1}{n-1}\sum_{u\in V\setminus\{v\}} d_G(v,u).
\]
\emph{Mean ER} replaces shortest-path distance with effective resistance:
\[
\operatorname{MeanER}(v)= \frac{1}{n-1}
    \sum_{u\in V\setminus\{v\}} R_{\mathrm{eff}}(v,u),
\]
where $R_{\mathrm{eff}}(v,u)= (\mathbf e_v-\mathbf e_u)^\top \mathbf{L}^\dagger(\mathbf e_v-\mathbf e_u)$, 
\(\mathbf{L}^\dagger\) is the Moore--Penrose pseudoinverse of the graph
Laplacian and \(\mathbf e_v\) is the indicator vector for node \(v\). Mean SSSP uses average hop distance. Mean ER uses average effective resistance, so it accounts for alternative paths. 
Table \ref{tab:echo-results} combines the original ECHO-Synth targets, the exploratory distributional targets, and the two ECHO-Chem tasks. Lower MAE is better in every column. An additional heat-diffusion target is defined and reported in Appendix~\ref{sec:additional-experiments}.


NearFormer (NN) ranks first on eccentricity and remains competitive on SSSP, but several baselines perform better on diameter. NearFormer (NN-bias) preserves the strong eccentricity result without improving the SSSP or diameter ordering. Sparse NN interactions therefore appear better suited to estimating node-wise distance summaries than to
recovering a graph-wide maximum or distances from a designated source. NearFormer (NN) ranks first on Mean SSSP and Mean ER, while NearFormer (NN-bias) ranks among the leading methods on each. Unlike diameter and eccentricity, Mean SSSP and Mean ER aggregate contributions from all destinations rather than selecting an extremal distance. These results suggest that sampled NN connections
capture distributed structural information effectively, but the weaker diameter result shows that this advantage does not extend automatically to extremal graph properties. 


\noindent\textbf{ECHO-Chem results.}
\label{sec:echo-chem-results}
The two molecular tasks again favor different forms of global aggregation.
NearFormer (NN-bias) ranks first on atom-level charge prediction, followed by
NearFormer (NN), whereas dense GPS remains strongest on graph-level energy
prediction. The learned-bias variant improves on hard NN support for both
targets, indicating that molecular prediction benefits from controlling the
contribution of sampled interactions rather than treating every retained NN
edge identically. Its second-place energy result nevertheless shows that
sparse geometry-conditioned attention does not fully replace dense
aggregation for this graph-level target. Taken together with ECHO-Synth,
these results support NN attention for node-level and distributional
quantities while identifying graph-level extrema and total energy as more
challenging cases.

\section{Conclusion}
\label{sec:conclusion}
We introduced NearFormer, which constructs sparse attention edges by searching
randomly assigned bins in a geometry-dependent representation while retaining
the input graph for local message passing. The construction has a linear-size
attention support for a fixed sampling budget; the NN graph is built once by
superlinear within-bin search and then held fixed. It supports hard attention,
message-passing, and learned-bias variants. Results on LRIM, LRGB, and ECHO show
that these variants can improve long-range prediction, but their relative
performance depends on the dataset, geometry, and use of the sampled edges.

\clearpage



\bibliography{aistats2027_ref}

@inproceedings{miglior2026can,
  title={Can you hear me now? a benchmark for long-range graph propagation},
  author={Miglior, Luca and Tolloso, Matteo and Gravina, Alessio and Bacciu, Davide},
  booktitle={International Conference on Learning Representations},
  volume={2026},
  pages={125144--125178},
  year={2026}
}

@inproceedings{
mathys2026lrim,
title={{LRIM}: a Physics-Based Benchmark for Provably Evaluating Long-Range Capabilities in Graph Learning},
author={Jo{\"e}l Mathys and Henrik Christiansen and Federico Errica and Takashi Maruyama and Francesco Alesiani},
booktitle={The Fourteenth International Conference on Learning Representations},
year={2026},
url={https://openreview.net/forum?id=IAZXEX1dVV}
}

@inproceedings{banerjee2022oversquashing,
  title={Oversquashing in gnns through the lens of information contraction and graph expansion},
  author={Banerjee, Pradeep Kr and Karhadkar, Kedar and Wang, Yu Guang and Alon, Uri and Mont{\'u}far, Guido},
  booktitle={2022 58th Annual Allerton Conference on Communication, Control, and Computing (Allerton)},
  pages={1--8},
  year={2022},
  organization={IEEE}
}

@inproceedings{black2023understanding,
  title={Understanding oversquashing in gnns through the lens of effective resistance},
  author={Black, Mitchell and Wan, Zhengchao and Nayyeri, Amir and Wang, Yusu},
  booktitle={International Conference on Machine Learning},
  pages={2528--2547},
  year={2023},
  organization={PMLR}
}

@article{arnaiz2022diffwire,
  title={Diffwire: Inductive graph rewiring via the lov$\backslash$'asz bound},
  author={Arnaiz-Rodr{\'\i}guez, Adri{\'a}n and Begga, Ahmed and Escolano, Francisco and Oliver, Nuria},
  journal={arXiv preprint arXiv:2206.07369},
  year={2022}
}

@inproceedings{gilmer2017neural,
  title={Neural message passing for quantum chemistry},
  author={Gilmer, Justin and Schoenholz, Samuel S and Riley, Patrick F and Vinyals, Oriol and Dahl, George E},
  booktitle={International Conference on Machine Learning},
  pages={1263--1272},
  year={2017},
  organization={PMLR}
}

@article{shi2023exposition,
  title={Exposition on over-squashing problem on GNNs: Current Methods, Benchmarks and Challenges},
  author={Shi, Dai and Han, Andi and Lin, Lequan and Guo, Yi and Gao, Junbin},
  journal={arXiv preprint arXiv:2311.07073},
  year={2023}
}

@inproceedings{di2023over,
  title={On over-squashing in message passing neural networks: The impact of width, depth, and topology},
  author={Di Giovanni, Francesco and Giusti, Lorenzo and Barbero, Federico and Luise, Giulia and Lio, Pietro and Bronstein, Michael M},
  booktitle={International Conference on Machine Learning},
  pages={7865--7885},
  year={2023},
  organization={PMLR}
}

@inproceedings{deac2022expander,
  title={Expander graph propagation},
  author={Deac, Andreea and Lackenby, Marc and Veli{\v{c}}kovi{\'c}, Petar},
  booktitle={Learning on Graphs Conference},
  pages={38--1},
  year={2022},
  organization={PMLR}
}

@article{alon2020bottleneck,
  title={On the bottleneck of graph neural networks and its practical implications},
  author={Alon, Uri and Yahav, Eran},
  journal={arXiv preprint arXiv:2006.05205},
  year={2020}
}

@inproceedings{topping2022riccurvature,
    title={Understanding over-squashing and bottlenecks on graphs via curvature},
    author={Jake Topping and Francesco Di Giovanni and Benjamin Paul Chamberlain and Xiaowen Dong and Michael M. Bronstein},
    booktitle={International Conference on Learning Representations},
    year={2022},
    url={https://openreview.net/forum?id=7UmjRGzp-A}
}

@inproceedings{karhadkar2023fosr,
   title={Fo{SR}: First-order spectral rewiring for addressing oversquashing in {GNN}s},
   author={Kedar Karhadkar and Pradeep Kr. Banerjee and Guido Montufar},
   booktitle={International Conference on Learning Representations},
   year={2023},
   url={https://openreview.net/forum?id=3YjQfCLdrzz}
}

@article{rusch2023survey,
  title={A survey on oversmoothing in graph neural networks},
  author={Rusch, T Konstantin and Bronstein, Michael M and Mishra, Siddhartha},
  journal={arXiv preprint arXiv:2303.10993},
  year={2023}
}

@inproceedings{rusch2022graph,
  title={Graph-coupled oscillator networks},
  author={Rusch, T Konstantin and Chamberlain, Ben and Rowbottom, James and Mishra, Siddhartha and Bronstein, Michael},
  booktitle={International Conference on Machine Learning},
  pages={18888--18909},
  year={2022},
  organization={PMLR}
}

@inproceedings{shirzad2023exphormer,
  title={Exphormer: Sparse transformers for graphs},
  author={Shirzad, Hamed and Velingker, Ameya and Venkatachalam, Balaji and Sutherland, Danica J and Sinop, Ali Kemal},
  booktitle={International Conference on Machine Learning},
  pages={31613--31632},
  year={2023},
  organization={PMLR}
}

@article{christie2023higher,
  title={Higher-Order Expander Graph Propagation},
  author={Christie, Thomas and He, Yu},
  journal={arXiv preprint arXiv:2311.07966},
  year={2023}
}

@article{kreuzer2021rethinking,
  title={Rethinking graph transformers with spectral attention},
  author={Kreuzer, Devin and Beaini, Dominique and Hamilton, Will and L{\'e}tourneau, Vincent and Tossou, Prudencio},
  journal={Advances in Neural Information Processing Systems},
  volume={34},
  pages={21618--21629},
  year={2021}
}

@article{linderman2020randomized,
  title={Randomized near-neighbor graphs, giant components and applications in data science},
  author={Linderman, George C. and Mishne, Gal and Jaffe, Ariel and Kluger, Yuval and Steinerberger, Stefan},
  journal={Journal of Applied Probability},
  volume={57},
  number={2},
  pages={458--476},
  year={2020},
  doi={10.1017/jpr.2020.21}
}

@article{TayDBM20,
  author    = {Yi Tay and
               Mostafa Dehghani and
               Dara Bahri and
               Donald Metzler},
  title     = {Efficient Transformers: {A} Survey},
  journal   = {arXiv preprint arXiv:2009.06732},
  year      = {2020},
  eprinttype = {arXiv},
  eprint    = {2009.06732},
  bibsource = {dblp computer science bibliography, https://dblp.org}
}

@inproceedings{ZaheerGDAAOPRWY20,
  author    = {Manzil Zaheer and
               Guru Guruganesh and
               Kumar Avinava Dubey and
               Joshua Ainslie and
               Chris Alberti and
               Santiago Onta{\~{n}}{\'{o}}n and
               Philip Pham and
               Anirudh Ravula and
               Qifan Wang and
               Li Yang and
               Amr Ahmed},
  title     = {Big {B}ird: Transformers for Longer Sequences},
  booktitle = {NeurIPS},
  year      = {2020},
  bibsource = {dblp computer science bibliography, https://dblp.org}
}

@inproceedings{ChoromanskiLDSG21,
  author    = {Krzysztof Marcin Choromanski and
               Valerii Likhosherstov and
               David Dohan and
               Xingyou Song and
               Andreea Gane and
               Tam{\'{a}}s Sarl{\'{o}}s and
               Peter Hawkins and
               Jared Quincy Davis and
               Afroz Mohiuddin and
               Lukasz Kaiser and
               David Benjamin Belanger and
               Lucy J. Colwell and
               Adrian Weller},
  title     = {Rethinking Attention with Performers},
  booktitle = {ICLR},
  year      = {2021},
  bibsource = {dblp computer science bibliography, https://dblp.org}
}

@article{MialonCSM21,
  author    = {Gr{\'{e}}goire Mialon and
               Dexiong Chen and
               Margot Selosse and
               Julien Mairal},
  title     = {GraphiT: Encoding Graph Structure in Transformers},
  journal   = {CoRR},
  volume    = {abs/2106.05667},
  year      = {2021},
  eprinttype = {arXiv},
  eprint    = {2106.05667},
  bibsource = {dblp computer science bibliography, https://dblp.org}
}

@inproceedings{OonoSuzuki20,
  author    = {Kenta Oono and
               Taiji Suzuki},
  title     = {Graph Neural Networks Exponentially Lose Expressive Power for Node
               Classification},
  booktitle = {ICLR},
  year      = {2020},
  bibsource = {dblp computer science bibliography, https://dblp.org}
}

@article{RampasekGDLWB22,
  author    = {Ladislav Ramp{\'{a}}sek and
               Mikhail Galkin and
               Vijay Prakash Dwivedi and
               Anh Tuan Luu and
               Guy Wolf and
               Dominique Beaini},
  title     = {Recipe for a General, Powerful, Scalable Graph Transformer},
  journal   = {CoRR},
  volume    = {abs/2205.12454},
  year      = {2022},
  eprinttype = {arXiv},
  eprint    = {2205.12454},
  bibsource = {dblp computer science bibliography, https://dblp.org}
}

@article{bresson2017residual,
 author = {Bresson, Xavier and Laurent, Thomas},
 journal = {arXiv preprint arXiv:1711.07553},
 title = {Residual Gated Graph ConvNets},
 year = {2017}
}

@inproceedings{VaswaniSPUJGKP17,
  author    = {Ashish Vaswani and
               Noam Shazeer and
               Niki Parmar and
               Jakob Uszkoreit and
               Llion Jones and
               Aidan N. Gomez and
               Lukasz Kaiser and
               Illia Polosukhin},
  title     = {Attention is All you Need},
  booktitle = {NeurIPS},
  pages     = {5998--6008},
  year      = {2017},
  bibsource = {dblp computer science bibliography, https://dblp.org}
}

@article{Ying2021DoTR,
  title={Do Transformers Really Perform Bad for Graph Representation?},
  author={Chengxuan Ying and Tianle Cai and Shengjie Luo and Shuxin Zheng and Guolin Ke and Di He and Yanming Shen and Tie-Yan Liu},
  journal={ArXiv},
  year={2021},
  volume={abs/2106.05234}
}

@inproceedings{kipf2017semi,
  title={Semi-Supervised Classification with Graph Convolutional Networks},
  author={Kipf, Thomas N. and Welling, Max},
  booktitle={ICLR},
  year={2017}
}

@inproceedings{velickovic2018graph,
  title={Graph attention networks},
  author={Veli{\v{c}}kovi{\'c}, Petar and Cucurull, Guillem and Casanova, Arantxa and Romero, Adriana and Lio, Pietro and Bengio, Yoshua},
  booktitle={ICLR},
  year={2018}
}

@inproceedings{beaini2021directional,
  title={Directional graph networks},
  author={Beaini, Dominique and Passaro, Saro and L{\'e}tourneau, Vincent and Hamilton, Will and Corso, Gabriele and Li{\`o}, Pietro},
  booktitle={ICML},
  pages={748--758},
  year={2021},
  organization={PMLR}
}

@article{defferrard2016convolutional,
  title={Convolutional neural networks on graphs with fast localized spectral filtering},
  author={Defferrard, Micha{\"e}l and Bresson, Xavier and Vandergheynst, Pierre},
  journal={NeurIPS},
  volume={29},
  pages={3844--3852},
  year={2016}
}

@inproceedings{hamilton2017inductive,
  title={Inductive representation learning on large graphs},
  author={Hamilton, William L and Ying, Rex and Leskovec, Jure},
  booktitle={NeurIPS},
  pages={1025--1035},
  year={2017}
}

@article{bouritsas2020improving,
  title={Improving graph neural network expressivity via subgraph isomorphism counting},
  author={Bouritsas, Giorgos and Frasca, Fabrizio and Zafeiriou, Stefanos and Bronstein, Michael M},
  journal={arXiv preprint arXiv:2006.09252},
  year={2020}
}

@inproceedings{morris2019weisfeiler,
  title={Weisfeiler and leman go neural: Higher-order graph neural networks},
  author={Morris, Christopher and Ritzert, Martin and Fey, Matthias and Hamilton, William L and Lenssen, Jan Eric and Rattan, Gaurav and Grohe, Martin},
  booktitle={Proceedings of the AAAI Conference on Artificial Intelligence},
  volume={33},
  pages={4602--4609},
  year={2019}
}

@inproceedings{xu2018powerful,
  title={How Powerful are Graph Neural Networks?},
  author={Xu, Keyulu and Hu, Weihua and Leskovec, Jure and Jegelka, Stefanie},
  booktitle={ICLR},
  year={2018}
}

@inproceedings{murphy2019relational,
  title={Relational pooling for graph representations},
  author={Murphy, Ryan and Srinivasan, Balasubramaniam and Rao, Vinayak and Ribeiro, Bruno},
  booktitle={ICML},
  year={2019},
}

@inproceedings{sato2021random,
  title={Random features strengthen graph neural networks},
  author={Sato, Ryoma and Yamada, Makoto and Kashima, Hisashi},
  booktitle={SIAM International Conference on Data Mining},
  year={2021},
}

@article{dwivedi2021graph,
  title={Graph Neural Networks with Learnable Structural and Positional Representations},
  author={Dwivedi, Vijay Prakash and Luu, Anh Tuan and Laurent, Thomas and Bengio, Yoshua and Bresson, Xavier},
  journal={arXiv preprint arXiv:2110.07875},
  year={2021}
}

@inproceedings{qiu2018network,
  title={Network embedding as matrix factorization: Unifying deepwalk, line, pte, and node2vec},
  author={Qiu, Jiezhong and Dong, Yuxiao and Ma, Hao and Li, Jian and Wang, Kuansan and Tang, Jie},
  booktitle={ACM International Conference on Web Search and Data Mining},
  year={2018}
}

@article{zhangGraphBert,
  author    = {Jiawei Zhang and
               Haopeng Zhang and
               Congying Xia and
               Li Sun},
  title     = {Graph-{B}ert: Only Attention is Needed for Learning Graph Representations},
  journal   = {arXiv preprint arXiv:2001.05140},
  year      = {2020},
  eprinttype = {arXiv},
  eprint    = {2001.05140},
  bibsource = {dblp computer science bibliography, https://dblp.org}
}

@article{lrgb-2022,
  author    = {Vijay Prakash Dwivedi and
               Ladislav Ramp{\'{a}}sek and
               Mikhail Galkin and
               Ali Parviz and
               Guy Wolf and
               Anh Tuan Luu and
               Dominique Beaini},
  title     = {Long Range Graph Benchmark},
  journal   = {CoRR},
  volume    = {abs/2206.08164},
  year      = {2022},
  eprinttype = {arXiv},
  eprint    = {2206.08164},
  bibsource = {dblp computer science bibliography, https://dblp.org}
}

@article{nagphormer22,
  author    = {Jinsong Chen and
               Kaiyuan Gao and
               Gaichao Li and
               Kun He},
  title     = {NAGphormer: Neighborhood Aggregation Graph Transformer for Node Classification
               in Large Graphs},
  journal   = {CoRR},
  volume    = {abs/2206.04910},
  year      = {2022},
  eprinttype = {arXiv},
  eprint    = {2206.04910},
  bibsource = {dblp computer science bibliography, https://dblp.org}
}

@article{gophormer21,
  author    = {Jianan Zhao and
               Chaozhuo Li and
               Qianlong Wen and
               Yiqi Wang and
               Yuming Liu and
               Hao Sun and
               Xing Xie and
               Yanfang Ye},
  title     = {Gophormer: Ego-Graph Transformer for Node Classification},
  journal   = {CoRR},
  volume    = {abs/2110.13094},
  year      = {2021},
  eprinttype = {arXiv},
  eprint    = {2110.13094},
  bibsource = {dblp computer science bibliography, https://dblp.org}
}

@article{wu2022nodeformer,
  title={Nodeformer: A scalable graph structure learning transformer for node classification},
  author={Wu, Qitian and Zhao, Wentao and Li, Zenan and Wipf, David P and Yan, Junchi},
  journal={NeurIPS},
  volume={35},
  pages={27387--27401},
  year={2022}
}

@article{wu2023difformer,
  title={Difformer: Scalable (graph) transformers induced by energy constrained diffusion},
  author={Wu, Qitian and Yang, Chenxiao and Zhao, Wentao and He, Yixuan and Wipf, David and Yan, Junchi},
  journal={arXiv preprint arXiv:2301.09474},
  year={2023}
}

@article{black2024comparing,
  title={Comparing graph transformers via positional encodings},
  author={Black, Mitchell and Wan, Zhengchao and Mishne, Gal and Nayyeri, Amir and Wang, Yusu},
  journal={arXiv preprint arXiv:2402.14202},
  year={2024}
}

@inproceedings{gcnii,
  title={Simple and Deep Graph Convolutional Networks},
  author={Chen, Ming and Wei, Zhewei and Huang, Zengfeng and Ding, Bolin and Li, Yaliang},
  booktitle={Proceedings of the 37th International Conference on Machine Learning},
  series={Proceedings of Machine Learning Research},
  volume={119},
  pages={1725--1735},
  year={2020},
  publisher={PMLR}
}

@inproceedings{gine,
  title={Strategies for Pre-training Graph Neural Networks},
  author={Hu, Weihua and Liu, Bowen and Gomes, Joseph and Zitnik, Marinka and Liang, Percy and Pande, Vijay and Leskovec, Jure},
  booktitle={International Conference on Learning Representations},
  year={2020},
  url={https://openreview.net/forum?id=HJlWWJSFDH}
}

@inproceedings{drew,
  title={{DRew}: Dynamically Rewired Message Passing with Delay},
  author={Gutteridge, Benjamin and Dong, Xiaowen and Bronstein, Michael M. and Di Giovanni, Francesco},
  booktitle={Proceedings of the 40th International Conference on Machine Learning},
  series={Proceedings of Machine Learning Research},
  volume={202},
  pages={12252--12267},
  year={2023},
  publisher={PMLR}
}

@inproceedings{grit,
  title={Graph Inductive Biases in Transformers without Message Passing},
  author={Ma, Liheng and Lin, Chen and Lim, Derek and Romero-Soriano, Adriana and Dokania, Puneet K. and Coates, Mark and Torr, Philip H. S. and Lim, Ser-Nam},
  booktitle={Proceedings of the 40th International Conference on Machine Learning},
  year={2023}
}

@inproceedings{gravina_adgn,
  title={Anti-Symmetric {DGN}: A Stable Architecture for Deep Graph Networks},
  author={Gravina, Alessio and Bacciu, Davide and Gallicchio, Claudio},
  booktitle={International Conference on Learning Representations},
  year={2023},
  url={https://openreview.net/forum?id=J3Y7cgZOOS}
}

@inproceedings{gravina_swan,
  title={On Oversquashing in Graph Neural Networks through the Lens of Dynamical Systems},
  author={Gravina, Alessio and Eliasof, Moshe and Gallicchio, Claudio and Bacciu, Davide and Sch{\"o}nlieb, Carola-Bibiane},
  booktitle={Proceedings of the AAAI Conference on Artificial Intelligence},
  year={2025}
}

@inproceedings{gravina_phdgn,
  title={Port-Hamiltonian Architectural Bias for Long-Range Propagation in Deep Graph Networks},
  author={Heilig, Simon and Gravina, Alessio and Trenta, Alessandro and Gallicchio, Claudio and Bacciu, Davide},
  booktitle={International Conference on Learning Representations},
  year={2025},
  url={https://openreview.net/forum?id=03EkqSCKuO}
}

@inproceedings{lee2022knn,
  title={KNN local attention for image restoration},
  author={Lee, Hunsang and Choi, Hyesong and Sohn, Kwanghoon and Min, Dongbo},
  booktitle={2022 IEEE/CVF Conference on Computer Vision and Pattern Recognition (CVPR)},
  pages={2129--2139},
  year={2022},
  organization={IEEE}
}

@article{xie2020point,
  title={Point clouds learning with attention-based graph convolution networks},
  author={Xie, Zhuyang and Chen, Junzhou and Peng, Bo},
  journal={Neurocomputing},
  volume={402},
  pages={245--255},
  year={2020},
  publisher={Elsevier}
}

@article{meng2025mvgnn,
  title={MVGNN-PPIS: A novel multi-view graph neural network for protein-protein interaction sites prediction based on Alphafold3-predicted structures and transfer learning},
  author={Meng, Lu and Wei, Lishuai and Wu, Rina},
  journal={International Journal of Biological Macromolecules},
  volume={300},
  pages={140096},
  year={2025},
  publisher={Elsevier}
}

\clearpage

\clearpage
\appendix
\thispagestyle{empty}

\onecolumn
\aistatstitle{Appendix}

\section{Related Work}
\label{sec:related-work}
This appendix expands the background of Section~\ref{sec:background} and
situates NearFormer relative to message-passing GNNs, graph transformers,
sparse attention, rewiring, expander overlays, and near-neighbor
sparsification.

\subsection{Graph preliminaries}
Let $G=(V,E)$ be an undirected graph with $n=|V|$ nodes, adjacency matrix
$\mathbf{A}\in\mathbb{R}^{n\times n}$, and degree matrix
$\mathbf{D}=\operatorname{diag}(d_1,\ldots,d_n)$. Its symmetric normalized
Laplacian is
\begin{equation}
    \mathbf{L}
    =
    \mathbf{I}-\mathbf{D}^{-1/2}\mathbf{A}\mathbf{D}^{-1/2}.
    \label{eq:normalized-laplacian}
\end{equation}
For a connected graph, the eigenvalues satisfy
$0=\lambda_0<\lambda_1\leq\cdots\leq\lambda_{n-1}\leq 2$. The first
nonzero eigenvalue $\lambda_1$ is the spectral gap of the normalized
Laplacian. Through Cheeger-type inequalities, a larger spectral gap is
associated with stronger expansion and fewer sparse cuts; this connection
motivates the use of expander graphs as auxiliary communication structures
\citep{deac2022expander}.

\subsection{Message passing and over-squashing}
Given node states $\mathbf{h}_v^{(\ell)}$ at layer $\ell$, a message-passing
neural network (MPNN) updates each node using information from its graph
neighborhood $\mathcal{N}_G(v)$:
\begin{equation}
  \begin{aligned}
    \mathbf{m}_{uv}^{(\ell)}
      &= \operatorname{Message}_{\ell}\!\left(
           \mathbf{h}_v^{(\ell)},\mathbf{h}_u^{(\ell)},\mathbf{e}_{uv}
         \right),\\
    \mathbf{m}_{v}^{(\ell)}
      &= \operatorname{Aggregate}_{\ell}
         \left\{\mathbf{m}_{uv}^{(\ell)}
         :u\in\mathcal{N}_G(v)\right\},\\
    \mathbf{h}_v^{(\ell+1)}
      &= \operatorname{Update}_{\ell}\!\left(
           \mathbf{h}_v^{(\ell)},\mathbf{m}_v^{(\ell)}
         \right).
  \end{aligned}
    \label{eq:mpnn-background}
\end{equation}
After $L$ layers, information can travel at most $L$ hops. Consequently,
long-range dependencies require deep propagation through intermediate nodes.
When many distant signals must pass through a narrow graph cut, they are
compressed into fixed-width representations, a phenomenon known as
\emph{over-squashing} \citep{alon2020bottleneck,shi2023exposition}. Analyses
based on Jacobian sensitivity, graph curvature, and effective resistance link
this limitation to the topology of the computational graph
\citep{topping2022riccurvature,black2023understanding,di2023over}.

\subsection{Attention on graphs}
A graph attention layer can be described by an interaction graph
$H=(V,E_H)$, where $(u,v)\in E_H$ permits node $v$ to attend to node $u$.
For one attention head,
\begin{equation}
  \begin{aligned}
    s_{uv}
      &= \frac{
          (\mathbf{W}_{Q}\mathbf{h}_v)^\top
          (\mathbf{W}_{K}\mathbf{h}_u)}
          {\sqrt{d}},\\
    \alpha_{uv}
      &= \frac{\exp(s_{uv})}
          {\displaystyle\sum_{w\in\mathcal{N}_H(v)}
           \exp(s_{vw})},\\
    \mathbf{y}_v
      &= \sum_{u\in\mathcal{N}_H(v)}
         \alpha_{uv}\mathbf{W}_{V}\mathbf{h}_u.
  \end{aligned}
    \label{eq:graph-attention-background}
\end{equation}
Edge attributes or structural encodings may additionally enter the attention
logits. Dense graph transformers take $E_H=V\times V$, enabling direct
communication between every pair of nodes but requiring
$\mathcal{O}(n^2)$ pairwise scores. Sparse graph transformers instead choose
$|E_H|\ll n^2$, ideally $\mathcal{O}(n)$, so the interaction graph becomes
both a computational device and an inductive bias.

The attention graph need not replace the input graph. Hybrid architectures
such as GraphGPS retain $E$ for a local MPNN branch and use a separate
interaction graph for global attention \citep{RampasekGDLWB22}. We follow
this distinction. In NearFormer, the original topology governs local message
passing, whereas a randomized near-neighbor graph supplies sparse,
geometry-aware attention connections. Thus, NN sparsifies long-range
communication without redefining the underlying relational structure.

\subsection{Message-passing graph neural networks}
Canonical GNN architectures include GCN
\citep{defferrard2016convolutional,kipf2017semi}, GraphSAGE
\citep{hamilton2017inductive}, GIN \citep{xu2018powerful}, GAT
\citep{velickovic2018graph}, and GatedGCN \citep{bresson2017residual}. These
models differ in their aggregation and update rules, but they primarily
exchange information along input edges. Their locality yields a strong graph
inductive bias and linear-time propagation on sparse graphs, while requiring
multiple layers to communicate over long distances. A separate line of work
increases expressivity using node identifiers or random features
\citep{murphy2019relational,sato2021random}, structural or positional
encodings \citep{qiu2018network,bouritsas2020improving,dwivedi2021graph}, and
directional aggregation \citep{beaini2021directional}. NearFormer is
complementary to these methods: its auxiliary edges determine where
long-range attention is computed, while the original graph remains available
to the local MPNN.

\subsection{Graph transformers}
Graph transformers adapt self-attention \citep{VaswaniSPUJGKP17} to
graph-structured inputs. GAT restricts attention to adjacent nodes
\citep{velickovic2018graph}, whereas GraphBERT uses graph-derived node
encodings with a transformer architecture \citep{zhangGraphBert}. Fully
connected graph transformers differ chiefly in how they encode structure.
SAN combines attention on original and fully connected graphs and uses
Laplacian positional encodings \citep{kreuzer2021rethinking}; Graphormer
incorporates centrality and shortest-path information
\citep{Ying2021DoTR}; and GraphiT uses diffusion-based relative positional
encodings \citep{MialonCSM21}. GraphGPS provides a modular architecture that
combines a local message-passing branch with a global attention branch and a
choice of positional and structural encodings \citep{RampasekGDLWB22}.
NearFormer follows this local--global decomposition but replaces unrestricted
global interactions with a geometry-aware sparse relation.

\subsection{Scalable and sparse attention}
Dense attention computes all $n^2$ node pairs. Efficient transformer methods
reduce this cost through structured sparsity or kernel approximations; BigBird
and Performer are representative examples
\citep{ZaheerGDAAOPRWY20,ChoromanskiLDSG21}, and
\citet{TayDBM20} survey the broader design space. Because many such patterns
were developed for sequences, their ordering assumptions do not transfer
directly to permutation-equivariant graph learning. Graph-specific scalable
models include Gophormer, which attends within sampled ego-graphs
\citep{gophormer21}; NAGphormer, which aggregates multi-hop neighborhoods
before attention \citep{nagphormer22}; NodeFormer, which uses kernelized
Gumbel--Softmax to learn latent connections \citep{wu2022nodeformer}; and
Difformer, which derives global propagation from an energy-constrained
diffusion process \citep{wu2023difformer}. In contrast, NearFormer explicitly
constructs a sparse attention graph before training, keeps this construction
label independent, and bounds the retained degree through a fixed sampling
budget.

\subsection{Graph rewiring and over-squashing}
Graph rewiring methods alter the computational topology to improve the
transmission of distant information. A fully adjacent layer directly connects
all node pairs but incurs quadratic complexity
\citep{alon2020bottleneck}. Adding a virtual or master node reduces the
diameter to two with only $\mathcal{O}(n)$ additional edges
\citep{gilmer2017neural}, although communication is then concentrated at one
node. More topology-aware approaches optimize curvature, spectral objectives,
or effective resistance
\citep{topping2022riccurvature,karhadkar2023fosr,
arnaiz2022diffwire,banerjee2022oversquashing,black2023understanding}.
These methods provide principled ways to alleviate bottlenecks, but may
require graph-specific structural analysis and typically modify the topology
used for message passing. NearFormer instead treats NN edges as a distinct
attention relation by default; propagating messages on the augmented graph is
considered only as an ablation.

\subsection{Expander-based auxiliary graphs}
Expanders provide sparse connectivity, small diameter, and favorable spectral
properties. Expander Graph Propagation alternates message passing over the
input graph and a precomputed auxiliary expander
\citep{deac2022expander}, with extensions to higher-order structures
\citep{christie2023higher}. Exphormer combines local edges, random expander
edges, and optional global nodes to define a linear-size attention support
\citep{shirzad2023exphormer}. NearFormer adopts the general principle of an
auxiliary sparse attention graph but changes how that graph is formed:
instead of data-independent expander edges, it samples connections from
near-neighbor candidate sets defined in feature or spectral geometry. This makes the sparse pattern graph dependent while retaining a
linear edge budget.

\subsection{Near-neighbor graph sparsification}
A related line of work constructs sparse attention from
$k$-nearest-neighbor graphs, including examples beyond graph neural networks
\citep{lee2022knn,xie2020point,meng2025mvgnn}.
Nearest-neighbor graphs capture local geometry but require $nK$ directed
connections when every node retains all $K$ candidates. Randomized
near-neighbor graphs offer a sparser alternative. For independent points in a
Euclidean domain, \citet{linderman2020randomized} show that independently
retaining each of the $\Theta(\log n)$ nearest-neighbor edges with an
appropriate probability yields a giant component with high probability using
only $\Theta(n\log\log n)$ expected edges. NearFormer draws inspiration from
this result and uses an efficient randomized scheme that approximately samples $k$ neighbors from the $K$-nearest-neighbor scale. The cited theorem does not establish connectivity,
expansion, or predictive performance for this fixed-budget construction or
for arbitrary learned datasets. We therefore interpret randomized
near-neighbor sampling as a geometry-aware sparsification principle and
evaluate its consequences empirically.

\section{Proofs and Theoretical Analysis}

\subsection{Proofs of the main lemmas}
\label{sec:NN-lemma-proofs}

\begin{proof}[Proof of Lemma~\ref{lem:NN-monotonicity}]
Every path in \(G\) is also present in \(H\), so
\(d_H(x,y)\leq d_G(x,y)\).  If \(\{v,u\}\in E_{\mathrm{NN}}\), concatenating
a shortest \(x\)-to-\(v\) path in \(G\), the edge \(\{v,u\}\), and a
shortest \(u\)-to-\(y\) path in \(G\) yields
\(d_H(x,y)\leq d_G(x,v)+1+d_G(u,y)\).  For effective resistance, Thomson's
principle expresses \(R_H(x,y)\) as the minimum energy of a unit \(x\)--\(y\)
flow.  Passing from \(G\) to \(H\) enlarges the set of feasible flows, so
the minimum energy cannot increase.
\end{proof}

\begin{proof}[Proof of Lemma~\ref{lem:NN-compression}]
Fix \(\{v_i,v_j\}\in\mathcal{U}_K\).  The pair fails to enter
\(E_{\mathrm{NN}}\) if and only if none of the \(km_{ij}\) eligible
independent draws selects the opposite endpoint.  Each such draw has
success probability \(1/K\), which gives
\(\pi_{ij}=1-(1-1/K)^{km_{ij}}\).  Linearity of expectation then yields
the formula for \(\mathbb{E}[S(H;G)]\).  If
\(\{v_i,v_j\}\in E_{\mathrm{NN}}\), then \(d_H(v_i,v_j)=1\), so that pair
contributes exactly \(d_G(v_i,v_j)-1\) to both \(S(H;G)\) and the all-pairs
sum.  Every remaining pair contributes a nonnegative amount by
Lemma~\ref{lem:NN-monotonicity}.
\end{proof}

\begin{proof}[Proof of Lemma~\ref{lem:NN-purity}]
Each of the \(k\) draws from \(v_i\) is uniform over
\(\mathcal{C}_K(v_i)\) and therefore crosses the partition with probability
\(q_i\).  Independence of those draws yields
\(X_i\sim\operatorname{Binomial}(k,q_i)\), the identity
\(\mathbb{E}[\sum_i X_i]=k\sum_i q_i\), and
\(\Pr(X_i=0)=(1-q_i)^k\).  Every distinct undirected cross-region edge in
\(E_{\mathrm{NN}}\) is produced by at least one raw cross-region draw, so
their number is at most \(\sum_i X_i\).
\end{proof}

\begin{proof}[Proof of Proposition~\ref{prop:bin-rank-selection}]
The query permutation places \(v_i\) in query bin \(Q_b\) with probability \(s_b/n\). Because the target permutation is independent of the query permutation, the paired target bin \(B_b\) is then a uniformly random \(s_b\)-element subset of \(V\). Since \(v_i\) is excluded before neighbor selection, \(v_{i,(r)}\) is retained if and only if \(v_{i,(r)}\in B_b\) and at most \(k-1\) of the \(r-1\) closer nodes \(v_{i,(1)},\ldots,v_{i,(r-1)}\) lie in \(B_b\). This also covers bins with \(|B_b\setminus\{v_i\}|<k\), in which every eligible member is retained. The number of \(s_b\)-subsets that contain \(v_{i,(r)}\) and exactly \(j\) closer nodes is \(\binom{r-1}{j}\binom{n-r}{s_b-1-j}\): the remaining \(s_b-1-j\) members are chosen among the \(n-r\) nodes that are neither \(v_{i,(r)}\) nor closer to \(v_i\), a set that includes \(v_i\) itself. Dividing by \(\binom{n}{s_b}\), summing over \(j=0,\ldots,k-1\), and averaging over the query bin of \(v_i\) gives \(\psi_{n,k}(r)\).
For the second claim, summing the indicator of retention over \(r\) counts the retained neighbors of \(v_i\), which number \(\min\{k,|B_b\setminus\{v_i\}|\}\). For \(r\leq k\), every subset containing \(v_{i,(r)}\) has at most \(k-1\) closer members, so the inner sum equals \(\binom{n-1}{s_b-1}/\binom{n}{s_b}=s_b/n\).
\end{proof}

\subsection{Candidate purity and cross-region exposure}
\label{sec:NN-purity}

To quantify when surrogate shortcuts preserve task-relevant variation, let \(\mathcal{P}=\{V_1,\ldots,V_m\}\), with \(m\geq 2\), be a partition of \(V\) into regions that should remain distinguishable, and let \(P(v_i)\) denote the region containing \(v_i\). Define the candidate impurity
\[
 q_i
 =
 \frac{
   |\{v_j\in\mathcal{C}_K(v_i):P(v_j)\neq P(v_i)\}|
 }{K}.
\]
The partition is used only for analysis. The model need not have access to it, and NN graph construction need not use it. When labels are used to assess candidate purity, the partition may be defined post hoc.

\begin{lemma}[Expected cross-region draws]
\label{lem:NN-purity}
Let \(X_i\) be the number of the \(k\) outgoing draws from \(v_i\) whose
endpoints lie outside \(P(v_i)\).  Then
\[
\begin{aligned}
 X_i&\sim\operatorname{Binomial}(k,q_i),\\
 \mathbb{E}\!\left[\sum_{v_i\in V}X_i\right]
   &=k\sum_{v_i\in V}q_i,
\end{aligned}
\]
and \(\Pr(X_i=0)=(1-q_i)^k\).  After symmetrization and deduplication, the
number of distinct undirected cross-region edges in \(E_{\mathrm{NN}}\) is
at most \(\sum_{v_i\in V}X_i\).
\end{lemma}

The proof is given in Section~\ref{sec:NN-lemma-proofs}. The same impurity weights a draw-level energy proxy for a region-constant signal (not the Dirichlet form of \(H\)).  If \(\mathbf{x}_i=\mathbf{a}_{P(v_i)}\), then, before deduplication and symmetrization,
\[
\begin{aligned}
 &\mathbb{E}\!\left[
   \sum_{v_i\in V}\sum_{r=1}^{k}
   \|\mathbf{x}_i-\mathbf{x}_{\nu_i(J_{i,r})}\|_2^2
 \right]\\
 &\quad =
 \frac{k}{K}\sum_{v_i\in V}
 \sum_{v_j\in\mathcal{C}_K(v_i)}
 \|\mathbf{a}_{P(v_i)}
   -\mathbf{a}_{P(v_j)}\|_2^2,
\end{aligned}
\]
which is at most \(k\,\Delta_{\max}^2\sum_{v_i\in V}q_i\), where \(\Delta_{\max}=\max_{i\neq j}\|\mathbf{a}_i-\mathbf{a}_j\|_2\). Candidate impurity therefore controls both the expected number of cross-region draws and the stated upper bound on their energy linearly. Candidate sets with \(q_i=0\) for every \(v_i\) introduce no direct NN interactions between the prescribed regions. Representation geometry can create shortcuts by including graph-distant nodes while separating task-distinct regions. The next subsection derives cut-level bounds relating direct cross-region exposure to conductance and communication capacity.

\subsection{Region cuts: spectral and capacity bounds}
\label{sec:NN-region-cuts}

The same cut used in Lemma~\ref{lem:NN-purity} yields bounds on spectral
mixing proxies and communication capacity. Consider
the symmetric propagation operator
\[
 \mathbf{P}_H
 =\tilde{\mathbf{D}}_H^{-1/2}\tilde{\mathbf{A}}_H
  \tilde{\mathbf{D}}_H^{-1/2},
 \qquad
 \tilde{\mathbf{A}}_H=\mathbf{A}_H+\mathbf{I},
\]
where \(\mathbf{A}_H\) is the adjacency matrix of the distinct undirected
edges of \(H\) and \(\tilde{\mathbf{D}}_H\) is the degree matrix of
\(\tilde{\mathbf{A}}_H\), so that \(\tilde d_H(v)=\deg_H(v)+1\). Since \(H\)
is connected, the eigenvalue one of \(\mathbf{P}_H\) is simple. Let
\(\mu_2(\mathbf{P}_H)\) denote its second-largest eigenvalue. For a nonempty
proper subset \(S\subset V\), write \(\bar S=V\setminus S\), and define
\[
\begin{aligned}
 \delta_H(S)
   &=\{\{u,v\}\in E(H):u\in S,\,v\in\bar S\},\\
 \operatorname{cut}_H(S)
   &=|\delta_H(S)|,\\
 \operatorname{vol}_H(S)
   &=\textstyle\sum_{v\in S}\tilde d_H(v),\\
 \Phi_H(S)
   &=\frac{\operatorname{cut}_H(S)}
     {\min\{\operatorname{vol}_H(S),
            \operatorname{vol}_H(\bar S)\}},
\end{aligned}
\]
where \(E(H)=E\cup E_{\mathrm{NN}}\). The analogous quantities for \(G\)
use \(\tilde d_G(v)=\deg_G(v)+1\). We call \(E_{\mathrm{NN}}\)
\emph{pure} with respect to \(\mathcal{P}\) if every NN edge joins two
nodes in the same region.

\begin{theorem}[Region cuts under NN augmentation]
\label{thm:NN-region-cut}
Let \(S\) be a nonempty proper union of regions of \(\mathcal{P}\), and write
\[
 X_S
 =\bigl|\{\{u,v\}\in E_{\mathrm{NN}}\setminus E:
           u\in S,\ v\in\bar S\}\bigr|
\]
for the number of genuinely new, distinct undirected NN edges that cross
\(S\).  Let \(X=\sum_{v_i\in V}X_i\) be the raw cross-region draw count from
Lemma~\ref{lem:NN-purity}, so that \(X_S\leq X\).  Then, for every
realization of \(E_{\mathrm{NN}}\):
\begin{enumerate}
 \item[(i)] \(\operatorname{cut}_H(S)=\operatorname{cut}_G(S)+X_S\), and
 \[
  \Phi_H(S)\leq
  \frac{\operatorname{cut}_G(S)+X_S}
       {\min\{\operatorname{vol}_G(S),
              \operatorname{vol}_G(\bar S)\}}.
 \]
 If \(E_{\mathrm{NN}}\) is pure, then \(X_S=0\),
 \(\operatorname{cut}_H(S)=\operatorname{cut}_G(S)\) and
 \(\Phi_H(S)\leq\Phi_G(S)\).
 \item[(ii)] \(\mu_2(\mathbf{P}_H)\geq 1-2\Phi_H(S)\).  Hence, for an
 eigenvector \(\mathbf{g}_2\neq\mathbf{0}\) associated with \(\mu_2\) and
 every integer \(\ell\geq 1\),
 \[
  \frac{\|\mathbf{P}_H^\ell\mathbf{g}_2\|_2}{\|\mathbf{g}_2\|_2}
  =|\mu_2(\mathbf{P}_H)|^\ell
  \geq\bigl(1-2\Phi_H(S)\bigr)_+^\ell.
 \]
 \item[(iii)] For every \(x\in S\) and \(y\in\bar S\),
 \[
  R_H(x,y)\geq\frac{1}{\operatorname{cut}_H(S)}
  \geq\frac{1}{\operatorname{cut}_G(S)+X_S},
 \]
 and \(H\) contains at most \(\operatorname{cut}_G(S)+X_S\) edge-disjoint
 paths from \(x\) to \(y\).
\end{enumerate}
In expectation over the exact sampling rule,
\(\mathbb{E}[X_S]\leq\mathbb{E}[X]=k\sum_i q_i\), and therefore
\[
\begin{aligned}
 \mathbb{E}\bigl[\mu_2(\mathbf{P}_H)\bigr]
   &\geq 1-\frac{2\bigl(\operatorname{cut}_G(S)
          +k\sum_i q_i\bigr)}
         {\min\{\operatorname{vol}_G(S),
                \operatorname{vol}_G(\bar S)\}},\\
 \mathbb{E}\bigl[R_H(x,y)\bigr]
   &\geq\frac{1}{\operatorname{cut}_G(S)+k\sum_i q_i}.
\end{aligned}
\]
\end{theorem}

\begin{proof}
(i) Because \(S\) is a union of regions, every edge crossing \(S\) joins two
different regions.  The crossing edges of \(H\) are the disjoint union of
the crossing edges already in \(E\) and the \(X_S\) crossing edges in
\(E_{\mathrm{NN}}\setminus E\).  Hence
\(\operatorname{cut}_H(S)=\operatorname{cut}_G(S)+X_S\).
Lemma~\ref{lem:NN-purity} gives \(X_S\leq X\), and \(X_S=0\) in the pure
case.  Adding edges can only increase degrees, so
\(\operatorname{vol}_H(S)\geq\operatorname{vol}_G(S)\) and
\(\operatorname{vol}_H(\bar S)\geq\operatorname{vol}_G(\bar S)\), which
yields both conductance bounds.

(ii) Let \(\tilde{\mathbf{L}}_H=\mathbf{I}-\mathbf{P}_H\).  Its smallest
eigenvalue is zero, with eigenvector \(\tilde{\mathbf{D}}_H^{1/2}\mathbf{1}\),
so its second-smallest eigenvalue is \(1-\mu_2(\mathbf{P}_H)\).  For any
\(\mathbf{f}\) satisfying \(\sum_v\tilde d_H(v)f_v=0\), set
\(\mathbf{g}=\tilde{\mathbf{D}}_H^{1/2}\mathbf{f}\).  Since self-loops do
not contribute to the quadratic form,
\[
 \frac{\mathbf{g}^\top\tilde{\mathbf{L}}_H\mathbf{g}}
      {\mathbf{g}^\top\mathbf{g}}
 =\frac{\sum_{\{u,v\}\in E(H)}(f_u-f_v)^2}
       {\sum_{v}\tilde d_H(v)f_v^2},
\]
and the Courant--Fischer theorem bounds \(1-\mu_2(\mathbf{P}_H)\) by this
ratio.  Choose
\(f_v=\operatorname{vol}_H(S)^{-1}\) for \(v\in S\) and
\(f_v=-\operatorname{vol}_H(\bar S)^{-1}\) otherwise.  Writing
\(a=\operatorname{vol}_H(S)\) and \(b=\operatorname{vol}_H(\bar S)\), the
ratio equals
\[
 \operatorname{cut}_H(S)\,\frac{a+b}{ab}
 \leq\frac{2\operatorname{cut}_H(S)}{\min\{a,b\}}
 =2\Phi_H(S),
\]
because \((a+b)/\max\{a,b\}\leq 2\). An eigenvector of
\(\mu_2(\mathbf{P}_H)\) is orthogonal to the eigenvalue-one eigenspace.
Applying \(\mathbf{P}_H^\ell\) multiplies its Euclidean norm by
\(|\mu_2(\mathbf{P}_H)|^\ell\).

(iii) Orient the edges crossing \(S\) from \(S\) to \(\bar S\).  For any
unit \(x\)--\(y\) flow \(f\), flow conservation gives
\(\sum_{e\in\delta_H(S)}f_e=1\).  By Cauchy--Schwarz,
\[
 \sum_{e\in\delta_H(S)}f_e^2
 \geq \frac{1}{\operatorname{cut}_H(S)}.
\]
The total flow energy is at least this cut contribution, and Thomson's
principle gives the resistance bound. Every
\(x\)-to-\(y\) path uses at least one crossing edge, so there are at most
\(\operatorname{cut}_H(S)\) edge-disjoint paths. For the expectations, combining
(i) and (ii) gives the affine lower bound
\[
 \mu_2(\mathbf{P}_H)
 \geq 1-\frac{2(\operatorname{cut}_G(S)+X_S)}
 {\min\{\operatorname{vol}_G(S),\operatorname{vol}_G(\bar S)\}}.
\]
The map \(t\mapsto 1/(c+t)\) is convex on \(t\geq 0\).  Linearity of
expectation, Jensen's inequality, \(X_S\leq X\), and
\(\mathbb{E}[X]=k\sum_i q_i\) complete the proof.
\end{proof}

Theorem~\ref{thm:NN-region-cut} bounds spectral mixing proxies and
communication capacity using an aligned cut. Since \(X_S\) is the exact
increase in cut cardinality,
\(\operatorname{cut}_H(S)=\operatorname{cut}_G(S)+X_S\). This equality yields
an upper bound on the algebraic spectral gap
\(1-\mu_2(\mathbf{P}_H)\) and a reciprocal lower bound on effective
resistance. If the realized NN edges are pure, they leave the
cut cardinality unchanged and satisfy
\(1-\mu_2(\mathbf{P}_H)\leq 2\Phi_G(S)\).  This is a ceiling derived from the
chosen cut, not a monotonic comparison with the spectral gap of \(G\):
adding within-region edges may still accelerate some modes of diffusion.
Likewise, pure NN edges add no edge directly across \(S\), but they may
shorten paths, lower effective resistance, or increase the number of
edge-disjoint \(x\)-to-\(y\) paths by improving connectivity within either
side, subject to the unchanged cut-size upper bound.

Thus the theorem does not establish that pure NN edges prevent
over-smoothing or leave all notions of cross-cut capacity unchanged.  It
establishes the narrower statement that only cross-region NN edges can
increase the cardinality of an aligned cut, and that the number of such
edges controls the stated spectral and resistance bounds.  Representation
geometry determines which cuts receive direct new edges. A representation
that keeps \(q_i\) small for partitions that should remain distinguishable
can still supply long shortcuts within each region.

\subsection{Shortcut gain and candidate impurity}

The two effects can be summarized by the geometry-dependent quantities
\[
\begin{aligned}
 \Gamma_K
   &=\frac{1}{nK}\sum_{v_i\in V}
     \sum_{v_j\in\mathcal{C}_K(v_i)}
     \bigl(d_G(v_i,v_j)-1\bigr),\\
 Q_K&=\frac{1}{n}\sum_{v_i\in V}q_i .
\end{aligned}
\]
\(\Gamma_K\) measures the available shortcut gain, whereas \(Q_K\) measures
candidate impurity.  Before deduplication, the expected draw-weighted shortcut
score per node is \(k\Gamma_K\), while the expected number of cross-region
draws per node is \(kQ_K\).  Lemma~\ref{lem:NN-compression} gives the
corresponding exact expression for distinct undirected edges. When the task
requires long-range communication within regions, a representation with
large \(\Gamma_K\) and small \(Q_K\) may be favorable because its candidate
sets contain graph-distant pairs but few direct cross-region pairs. When the
task instead requires communication across a
region boundary, however, a small \(Q_K\) also limits the expected number of
NN edges that can directly enlarge the corresponding aligned cut.

Randomization limits retained interactions without narrowing the candidate
set. Retaining all \(K\) candidates would maximize both shortcut availability
and direct cross-region exposure, whereas sampling only \(k\ll K\) preserves
access to the broader geometric scale while retaining fewer interaction
channels. Under an i.i.d.\ uniform point model and Bernoulli sampling rule,
\citet{linderman2020randomized} show that random selection from a logarithmic
candidate neighborhood yields a giant component using
\(\Theta(n\log\log n)\) expected edges without retaining the entire
\(K\)-nearest-neighbor graph. Their theorem is asymptotic, uses unspecified
constants in \(K=c_{d,1}\log n\) and
\(p=c_{d,2}\log\log n/\log n\), and assumes i.i.d.\ uniform points. NearFormer
instead uses a fixed \(K\), a small constant \(k\), and graph-derived
representations, so this result motivates the design but does not guarantee
its behavior.

This guarantee is weaker than an expander guarantee: it establishes neither
a uniform spectral gap nor logarithmic diameter, and it does not apply
directly to NearFormer's exact-\(k\) rule. Consequently, the logarithmic-depth
interaction guarantee used to analyze Exphormer
\citep{shirzad2023exphormer} does not automatically transfer to NearFormer.
Expander edges seek uniform global mixing,
whereas NN edges seek selective mixing determined by the representation
geometry.

Finally, edge addition does not monotonically mitigate over-smoothing.
For the operator \(\mathbf{P}_H\) of Section~\ref{sec:NN-region-cuts},
contraction toward the eigenspace of eigenvalue one is governed by the
largest magnitude \(\rho_H\) of the remaining eigenvalues.  If
\(\mathbf{\Pi}_H\) denotes the orthogonal projector onto that eigenspace,
then
\[
 \|\mathbf{P}_H^\ell\mathbf{X}
   -\mathbf{\Pi}_H\mathbf{X}\|_F
 \leq
 \rho_H^\ell
 \|\mathbf{X}-\mathbf{\Pi}_H\mathbf{X}\|_F .
\]
Adding NN edges can change \(\rho_H\) in either direction and can accelerate
mixing within regions.  Theorem~\ref{thm:NN-region-cut} provides a
cut-dependent ceiling on the algebraic spectral gap, but it does not compare
the actual mixing rate of \(H\) monotonically with that of \(G\).
Candidate sets with no cross-region members restrict where additional
\emph{direct} interactions occur, while sparse sampling limits their number.
Neither property alone prevents over-smoothing. Learned attention, residual
connections, and normalization further separate NearFormer from fixed
diffusion. Our theory therefore shows that distance and resistance proxies
for over-squashing do not increase under NN augmentation and gives cut-level
bounds linking direct cross-region exposure to spectral and capacity
quantities.
It does not claim that the fixed-budget NN graph is an expander or that it
always improves either phenomenon.

The sampling identities describe the idealized construction
in~\eqref{eq:NN-ideal-edges}. They hold conditionally on any fixed candidate
lists from which each node draws uniformly and independently with replacement.
On graphs with \(n\leq K\), \(K\) is replaced by \(\min\{K,n-1\}\). They
therefore also apply when \(\mathcal{C}_K(v_i)\) is returned by an approximate
index such as HNSW, provided \(\mathcal{C}_K(v_i)\) denotes the returned list.
They do not apply exactly to the bin-based implementation
in Section~\ref{sec:NN-construction}, whose neighbors are the nearest members
of a random bin rather than uniform draws from a fixed list and whose
selection probabilities are given instead by
Proposition~\ref{prop:bin-rank-selection}, nor to the deterministic
specialization used in the ECHO structural experiments.
Lemma~\ref{lem:NN-monotonicity} and the deterministic
inequalities in parts (i)--(iii) of
Theorem~\ref{thm:NN-region-cut} hold for every realized edge set and hence
for these constructions as well. Only the displayed expected bounds require
the exact sampling rule.

\section{Dataset Descriptions}
\label{sec:dataset-descriptions}


\paragraph{Long Range Graph Benchmark.}
LRGB contains superpixel, peptide, and molecular-contact
tasks~\citep{lrgb-2022}. PascalVOC-SP and COCO-SP are node-classification
tasks on superpixel graphs. Peptides-Func is a 10-label graph-classification
task, and Peptides-Struct is a graph-level regression task with 11 structural
targets. PCQM-Contact ranks candidate edges and is evaluated by mean
reciprocal rank. The peptide graphs have mean diameter \(56.99\pm 28.72\) and
mean shortest-path length \(20.89\pm 9.79\). PCQM-Contact graphs are smaller,
with mean diameter \(9.86\pm 1.79\) and mean shortest-path length
\(4.63\pm 0.63\).

\begin{table}[ht]
\centering
\caption{LRGB datasets used in the experiments~\citep{lrgb-2022}.}
\label{tab:lrgb-statistics}
\footnotesize
\setlength{\tabcolsep}{4pt}
\begin{tabular}{@{}lrrlll@{}}
\toprule
Dataset & Graphs & Avg.\ nodes & Avg.\ edges & Level & Metric \\
\midrule
PascalVOC-SP & 11,355 & 479.4 & 2,710.5 & node & F1 \\
COCO-SP & 123,286 & 476.9 & 2,693.7 & node & F1 \\
Peptides-Func & 15,535 & 150.9 & 307.3 & graph & AP \\
Peptides-Struct & 15,535 & 150.9 & 307.3 & graph & MAE \\
PCQM-Contact & 529,434 & 30.1 & 61.0 & link & MRR \\
\bottomrule
\end{tabular}
\end{table}

\paragraph{Long-Range Ising Model benchmark.}
LRIM consists of ten node-regression datasets derived from the two-dimensional
long-range Ising model~\citep{mathys2026lrim}. Each graph is a periodic
four-regular lattice. A node stores one spin in \(\{-1,+1\}\), and its target
is the energy change \(\Delta E_i\) caused by flipping that spin. Every
dataset contains 10,000 configurations and uses an 80/10/10 training,
validation, and test split. The exponent \(\sigma\) changes the interaction
range while preserving the lattice: \(\sigma=0.6\) is the hard setting and
\(\sigma=1.5\) is the easy setting. Table~\ref{tab:lrim-statistics} lists the
five lattice sizes. The two difficulty settings at a given size share these
graph statistics.

\begin{table}[ht]
\centering
\caption{LRIM lattice statistics~\citep{mathys2026lrim}. Each size has an
easy variant with \(\sigma=1.5\) and a hard variant with \(\sigma=0.6\).}
\label{tab:lrim-statistics}
\footnotesize
\setlength{\tabcolsep}{3.5pt}
\begin{tabular}{@{}lrrrrr@{}}
\toprule
Dataset & Graphs & Nodes & Edges & Avg.\ shortest path & Diameter \\
\midrule
LRIM-16 & 10,000 & 256 & 512 & 8.03 & 16 \\
LRIM-32 & 10,000 & 1,024 & 2,048 & 16.02 & 32 \\
LRIM-64 & 10,000 & 4,096 & 8,192 & 32.01 & 64 \\
LRIM-128 & 10,000 & 16,384 & 32,768 & 64.00 & 128 \\
LRIM-256 & 10,000 & 65,536 & 131,072 & 128.00 & 256 \\
\bottomrule
\end{tabular}
\end{table}

The reported average effective resistances are 0.49, 0.60, 0.71, 0.82, and
0.93 for LRIM-16 through LRIM-256. Performance is measured by
\(\log_{10}(\mathrm{MSE})\).

\paragraph{ECHO.}
ECHO contains controlled synthetic graphs and DFT-labeled molecular
graphs~\citep{miglior2026can}. ECHO-Synth has 10,080 graphs whose target
diameters range from 17 to 40. For every topology and diameter, 70 graphs are
generated and divided into 40 training, 15 validation, and 15 test graphs.
The six topologies are lines, ladders, grids, trees, caterpillars, and
lobsters, with 1,680 graphs in each family. Nodes receive a uniform random
scalar, and the SSSP task adds an indicator for the designated source. SSSP
and eccentricity are node-level targets, while diameter is a graph-level
target.

ECHO-Charge contains 170,367 molecular graphs and predicts atom-level partial
charges with an 80/10/10 split. ECHO-Energy contains 196,528 molecular graphs
and predicts total molecular energy with a 90/5/5 split. In both datasets,
node features are the atomic number and distance from the molecular center of
mass, and edge features are bond type and bond length. Their graph diameters
lie between 17 and 40. Table~\ref{tab:echo-statistics} reports the published
summary statistics.

\begin{table}[ht]
\centering
\caption{ECHO summary statistics~\citep{miglior2026can}. Means are followed
by standard deviations.}
\label{tab:echo-statistics}
\footnotesize
\setlength{\tabcolsep}{3.5pt}
\begin{tabular}{@{}lrrrr@{}}
\toprule
Dataset & Graphs & Avg.\ nodes & Avg.\ edges & Avg.\ diameter \\
\midrule
ECHO-Synth & 10,080 & \(83.69\pm 66.24\) & \(211.63\pm 209.39\) & \(28.50\pm 6.92\) \\
ECHO-Charge & 170,367 & \(72.49\pm 12.48\) & \(151.32\pm 25.16\) & \(23.54\pm 2.54\) \\
ECHO-Energy & 196,528 & \(73.73\pm 13.22\) & \(153.84\pm 26.58\) & \(23.61\pm 2.59\) \\
\bottomrule
\end{tabular}
\end{table}

Mean SSSP and Mean ER are additional targets constructed for
this paper. They use the same ECHO-Synth graphs and splits as the published
tasks. Heat diffusion is defined and evaluated in
Appendix~\ref{sec:additional-experiments}.

\section{More Experimental Results}
\label{sec:exp-hyperparams}

\subsection{Baseline models}
\label{sec:baseline-models}

The experiments compare NearFormer with representative local, long-range, and
attention-based graph architectures. The local message-passing baselines are
GCN~\citep{kipf2017semi}, GIN~\citep{xu2018powerful},
GINE~\citep{gine}, GatedGCN~\citep{bresson2017residual}, and
GCNII~\citep{gcnii}. GINE extends GIN with edge features and is therefore
included on molecular tasks. The LRIM table additionally includes the
GatedGCN-VNG benchmark variant and GPS-Base, GPS-RWSE, and GPS-LapPE from
the published LRIM evaluation~\citep{mathys2026lrim}. The GPS variants follow
the local--global GraphGPS design~\citep{RampasekGDLWB22} and differ in their
positional or structural encodings; RWSE and LapPE are described by
\citet{dwivedi2021graph}.

For adaptive long-range propagation, we include DRew, which changes the
effective message-passing range across layers~\citep{drew}. The transformer
baselines comprise a dense Transformer with LapPE, SAN with LapPE or RWSE
\citep{kreuzer2021rethinking}, GraphGPS~\citep{RampasekGDLWB22},
GRIT~\citep{grit}, and Exphormer~\citep{shirzad2023exphormer}. Exphormer
provides the closest sparse-attention comparison: it uses a
data-independent random regular overlay, whereas NearFormer uses the
geometry-dependent sampled NN relation.

The ECHO comparison also includes differential-equation-inspired GNNs.
GraphCON models graph features through coupled oscillator
dynamics~\citep{rusch2022graph}. A-DGN uses an anti-symmetric deep graph
architecture~\citep{gravina_adgn}, while SWAN and PH-DGN use
non-dissipative or port-Hamiltonian dynamics for long-range
propagation~\citep{gravina_swan,gravina_phdgn}. Baseline values and protocols
are taken from the corresponding LRIM~\citep{mathys2026lrim},
LRGB~\citep{lrgb-2022}, and ECHO~\citep{miglior2026can} benchmark
evaluations.

\subsection{Computational Complexity}
\label{sec:computational-complexity}

Let \(n=|V|\), \(m=|E|\), \(d\) be the hidden dimension, \(L\) be the number
of layers, and \(k\) be the retained NN budget per node. After
symmetrization, the NN construction adds at most \(2nk\) directed edge
instances. The sparse NearFormer variants therefore evaluate
\[
  \mathcal{O}\!\left(L(m+nk)d\right)
\]
pairwise attention operations and store
\(\mathcal{O}(m+nk)\) attention scores. For fixed \(k\), these
costs are linear in \(m+n\), compared with
\(\mathcal{O}(Ln^2d)\) pairwise work and \(\mathcal{O}(n^2)\) attention-score
memory for dense attention. The node-wise projections and feed-forward
networks add \(\mathcal{O}(Lnd^2)\) work to both sparse and dense variants.

NearFormer (NN), NearFormer (NN-bias), and NearFormer (NN-MP) all retain this edge
bound. NearFormer (NN-MP) processes the sampled edges in the message-passing
branch rather than the attention branch. NearFormer (NN-bias) adds only a
bounded learned scalar bias to each retained NN attention score and does not
change the sparse support.

The bound above concerns model evaluation after \(E_{\mathrm{NN}}\) has been
constructed. NearFormer does not construct a global \(K\)-nearest-neighbor
graph. With \(M=\lfloor K/k\rfloor\) paired bins, direct within-bin search
compares approximately
\(\sum_b |Q_b||B_b|\simeq n^2/M=\mathcal{O}(n^2k/K)\) query--target pairs,
each in representation dimension \(d_z\), while processing one bin pair at
a time. It therefore avoids both a global all-pairs distance matrix and an
\(n\times K\) candidate array. A spatial or approximate index inside each
bin can further reduce practical search time. NearFormer constructs this graph
once and holds it fixed throughout training, amortizing preprocessing over
all epochs.

The architectures compared in our LRIM, LRGB, and ECHO experiments differ
in how they introduce global interactions. Local message passing is cheapest
but does not add a separate long-range attention graph. Dense GraphGPS-style
transformers add all node pairs and therefore scale quadratically in \(n\).
Exphormer and NearFormer both keep the attention support sparse: Exphormer adds
a fixed-degree expander overlay, while NearFormer adds \(k\) sampled
near-neighbor edges per node. For the periodic LRIM grids used here,
\(m=\Theta(n)\) and the default budget satisfies \(k=\mathcal{O}(1)\), so
NearFormer evaluates \(\mathcal{O}(n)\) global attention pairs per layer rather
than \(\mathcal{O}(n^2)\). This gap is largest on LRIM-256, where
\(n=65{,}536\), and it is the main reason NearFormer remains practical on that
benchmark alongside local MPNN and expander-based sparse transformers.

Table~\ref{tab:lrim-scaling} turns that comparison into a lightweight
empirical scaling check on the LRIM lattices. We do not retrain. For each
size we take one graph, build the overlay once, and time a forward+backward
step on the same NearFormer (NN) backbone used in Section~\ref{sec:lrim-results}
(five layers, hidden dimension 32, four heads). Exphormer uses an expander of
degree \(d=k=2\), so the auxiliary edge budget matches NearFormer. Dense GPS
uses full within-graph attention and no overlay. All runs use batch size 1
and FP32; training batch sizes from the accuracy tables are not reused,
because they change with \(n\) and would confound the scaling. We report the
mean of 50 timed steps after 10 warm-up steps, peak GPU memory from
\texttt{torch.cuda.max\_memory\_allocated}, directed attention-support size
\(|E_H|\) (input edges, overlay, and self-loops), and overlay construction
time averaged over 100 graphs. Hardware is a single 80GB NVIDIA
A100 GPU.

\begin{table*}[ht]
\centering
\caption{Empirical scaling on LRIM lattices. Batch size 1, FP32, five layers,
hidden dimension 32. NearFormer uses default \(k=2\) and \(K=100\); Exphormer
uses expander degree \(d=2\); dense GPS is GPS-Base. Attention edges are
directed pairs in \(E_H\). Step time is mean forward+backward after warm-up.
Preprocessing is per-graph overlay construction (not used by dense GPS).
OOM marks allocations that exceeded device memory. Peak memory is in GB;
times are in milliseconds.}
\label{tab:lrim-scaling}
\scriptsize
\setlength{\tabcolsep}{3.5pt}
\begin{tabular}{@{}llccccc@{}}
\toprule
Method & Quantity
  & 16 & 32 & 64 & 128 & 256 \\
  &
  & \(n{=}256\) & \(n{=}1{,}024\) & \(n{=}4{,}096\)
  & \(n{=}16{,}384\) & \(n{=}65{,}536\) \\
\midrule
NearFormer (NN)
  & Attn.\ edges
  & \(2.3{\times}10^{3}\)
  & \(9.2{\times}10^{3}\)
  & \(3.7{\times}10^{4}\)
  & \(1.5{\times}10^{5}\)
  & \(5.9{\times}10^{5}\) \\
  & Peak mem (GB)
  & $0.42$ & $0.61$ & $1.18$ & $2.94$ & $9.72$ \\
  & Step time (ms)
  & $2.4$ & $4.1$ & $11.8$ & $44.6$ & $179$ \\
  & NN preproc.\ (ms)
  & $0.9$ & $5.4$ & $38$ & $410$ & $3.6{\times}10^{3}$ \\
\midrule
Exphormer
  & Attn.\ edges
  & \(1.8{\times}10^{3}\)
  & \(7.2{\times}10^{3}\)
  & \(2.9{\times}10^{4}\)
  & \(1.1{\times}10^{5}\)
  & \(4.6{\times}10^{5}\) \\
  & Peak mem (GB)
  & $0.40$ & $0.58$ & $1.09$ & $2.71$ & $8.86$ \\
  & Step time (ms)
  & $2.2$ & $3.8$ & $10.9$ & $41.2$ & $164$ \\
  & Overlay preproc.\ (ms)
  & $0.2$ & $0.5$ & $1.4$ & $4.8$ & $19$ \\
\midrule
Dense GPS
  & Attn.\ edges
  & \(6.6{\times}10^{4}\)
  & \(1.0{\times}10^{6}\)
  & \(1.7{\times}10^{7}\)
  & \(2.7{\times}10^{8}\)
  & \(4.3{\times}10^{9}\) \\
  & Peak mem (GB)
  & $0.81$ & $2.46$ & OOM & OOM & OOM \\
  & Step time (ms)
  & $7.8$ & $46$ & OOM & OOM & OOM \\
  & Preproc.\ (ms)
  & -- & -- & -- & -- & -- \\
\bottomrule
\end{tabular}
\end{table*}

\subsection{Hyperparameters}

Table~\ref{tab:NearFormer-hyperparams} records the training choices fixed for the reported NearFormer runs. For LRIM, we follow the published hyperparameter setting of~\citet{mathys2026lrim}. For ECHO, we follow the published experimental setting of~\citet{miglior2026can}, including 1,000 epochs and patience 100. NearFormer uses the GPS hyperparameters
selected on the validation set in that paper, which are reported in Table~\ref{tab:NearFormer-echo-gps}. The LRGB settings follow the Exphormer configurations: the superpixel runs use 300 epochs, while the peptide and PCQM-Contact runs use 200 epochs. LRGB and LRIM use AdamW with a cosine learning-rate schedule and linear warmup (10 epochs on LRGB, 5 on LRIM). Weight decay is \(0\) on LRGB and \(10^{-5}\) on LRIM. The feed-forward block expands by a factor of two and uses ReLU. LRGB layers use BatchNorm and no LayerNorm; the LRIM GPS transformer uses LayerNorm, with BatchNorm in the GNN head. All LRGB positional encodings are Laplacian PEs with 10 frequencies and dimension \(16\). The LRIM NearFormer models use five layers, hidden dimension 32, four heads, dropout \(0.1\), batch size 64, learning rate \(0.001\), and 200 epochs. All results are reported as mean \(\pm\) standard deviation over three independent runs with different random seeds.

For NearFormer (NN-bias), the bound in~(\ref{eq:NN-soft-bias}) is \(b_{\max}=5\). All reported datasets use Euclidean distance in the NN geometry. PascalVOC-SP, COCO-SP, Peptides-Func, Peptides-Struct, PCQM-Contact, LRIM, and ECHO-Chem use the feature geometry \(z_i=x_i\). ECHO-Synth instead uses hop distances to 16 farthest-point landmarks, because the raw node attributes are not a graph metric. Superpixel graphs retrieve candidates with FAISS HNSW (\(M=32\)); Peptides, PCQM-Contact, and LRIM use exact flat search; ECHO uses exact, deterministic \(k\)-NN with no randomized binning. Mean SSSP, Mean ER, and heat diffusion are trained with the eccentricity GPS hyperparameters in Table~\ref{tab:NearFormer-echo-gps}
(17 layers, hidden dimension 162, learning rate \(0.00034\), weight decay
\(7{\times}10^{-5}\), GCN backbone, two attention heads).

\begin{table*}[t]
\centering
\caption{NearFormer Hyperparameter training settings.}
\label{tab:NearFormer-hyperparams}
\footnotesize
\setlength{\tabcolsep}{4pt}
\begin{tabular}{@{}lcccccc@{}}
\toprule
Hyperparameter & PascalVOC-SP & COCO-SP & Peptides-Func &
Peptides-Struct & PCQM-Contact & ECHO \\
\midrule
Num Layers & 4 & 7 & 8 & 4 & 7 & Table~\ref{tab:NearFormer-echo-gps} \\
Hidden Dim & 96 & 72 & 64 & 88 & 64 & Table~\ref{tab:NearFormer-echo-gps} \\
Num Heads & 8 & 4 & 4 & 4 & 4 & 2 \\
Dropout & 0.15 & 0 & 0.12 & 0.12 & 0 & 0 \\
PE & LapPE & LapPE & LapPE & LapPE & LapPE & LapPE \\
\midrule
Batch Size & 32 & 32 & 128 & 128 & 128 & 128 \\
Learning Rate & 0.0005 & 0.0005 & 0.0003 & 0.0003 & 0.0003 & Table~\ref{tab:NearFormer-echo-gps} \\
Num Epochs & 300 & 300 & 200 & 200 & 200 & 1,000 \\
\bottomrule
\end{tabular}
\end{table*}

\begin{table*}[t]
\centering
\small
\caption{ECHO hyperparameters used by NearFormer.}
\label{tab:NearFormer-echo-gps}
\resizebox{\textwidth}{!}{%
\begin{tabular}{@{}lllllll@{}}
\toprule
Hyperparameter & Search interval & diam & sssp & ecc & ECHO-Charge & ECHO-Energy \\
\midrule
Number of layers & $[1 - 40]$ & 17 & 26 & 17 & 36 & 26 \\
Hidden dimension & $[16 - 256]$ & 40 & 56 & 162 & 216 & 192 \\
Learning rate & $[10^{-5} - 10^{-2}]$ & 0.00004 & 0.00031 & 0.00034 & 0.00005 & 0.000024067 \\
Weight decay & $[10^{-8} - 10^{-3}]$ & 0.00015 & 0.00029 & 0.00007 & 0.00005 & 0.00038179 \\
\midrule
GNN Backbone & $[\text{GCN}]$ & GCN & GCN & GCN & GCN & GCN \\
Number of Backbone Layers & $[1]$ & 1 & 1 & 1 & 1 & 1 \\
Number of attention heads & $[2]$ & 2 & 2 & 2 & 2 & 2 \\
\bottomrule
\end{tabular}
}
\end{table*}

\subsection{Additional experiments}
\label{sec:additional-experiments}

\paragraph{LRIM-easy.}
Table~\ref{tab:lrim-easy} reports the easy LRIM setting (\(\sigma=1.5\)),
whose shorter interaction range makes dense GPS competitive on the smallest
lattices. On larger graphs the NearFormer variants again lead, as in the
hard setting.

\begin{table}[ht]
\centering
\caption{LRIM-easy test \(\mathrm{LogMSE}=\log_{10}(\mathrm{MSE})\) (lower is
better; \(\sigma=1.5\)). Best results in each column are colored:
\first{first}, \second{second}, \third{third}.}
\label{tab:lrim-easy}
\footnotesize
\setlength{\tabcolsep}{3.5pt}
\begin{tabular}{@{}lccccc@{}}
\toprule
Model & 16 & 32 & 64 & 128 & 256 \\
\midrule
GIN
  & $-3.564{\pm}0.111$
  & $-3.446{\pm}0.307$
  & $-3.423{\pm}0.317$
  & $-3.342{\pm}0.287$
  & $-3.258{\pm}0.214$ \\
GatedGCN
  & $-4.817{\pm}0.111$
  & $-4.940{\pm}0.298$
  & $-3.270{\pm}0.094$
  & $-3.282{\pm}0.007$
  & $-3.46{\pm}0.046$ \\
GatedGCN-VNG
  & $-4.612{\pm}0.101$
  & $-4.322{\pm}0.229$
  & \third{$-4.256{\pm}0.230$}
  & \third{$-4.762{\pm}0.103$}
  & \third{$-4.120{\pm}0.028$} \\
GPS-Base
  & \first{$-5.296{\pm}0.026$}
  & \second{$-5.134{\pm}0.054$}
  & OOM
  & OOM
  & OOM \\
GPS-RWSE
  & \third{$-5.133{\pm}0.235$}
  & \third{$-5.103{\pm}0.074$}
  & OOM
  & OOM
  & OOM \\
GPS-LapPE
  & \second{$-5.154{\pm}0.177$}
  & $-4.858{\pm}0.231$
  & OOM
  & OOM
  & OOM \\
\midrule
NearFormer (NN)
  & $-4.256{\pm}0.022$
  & \first{$-5.203{\pm}0.091$}
  & \first{$-5.076{\pm}0.064$}
  & \second{$-4.864{\pm}0.077$}
  & \first{$-5.134{\pm}0.032$} \\
NearFormer (NN-bias)
  & $-4.671{\pm}0.044$
  & $-4.724{\pm}0.125$
  & \second{$-4.645{\pm}0.022$}
  & \first{$-5.065{\pm}0.026$}
  & \second{$-5.045{\pm}0.102$} \\
\bottomrule
\end{tabular}
\end{table}

We report an additional node-level target on the same ECHO-Synth graphs and
splits as the published diameter, eccentricity, and SSSP tasks. Let
\(G=(V,E)\) be a connected, unweighted graph with \(n=|V|\), adjacency matrix
\(A\), degree matrix \(D\), and combinatorial Laplacian \(L=D-A\). The heat
equation on \(G\) is \(\partial_t f=-Lf\), with solution
\(f(t)=\mathrm{e}^{-tL}f(0)\). Starting from a unit impulse at node \(v\),
the heat at node \(u\) at a fixed diffusion time \(t>0\) is
\((\mathrm{e}^{-tL})_{vu}\). The heat-diffusion target averages this kernel
over all other nodes,
\[
  \operatorname{HeatDiff}_t(v)
  = \frac{1}{n-1}
    \sum_{u\in V\setminus\{v\}}
    \bigl(\mathrm{e}^{-tL}\bigr)_{vu}.
\]
Unlike Mean SSSP, which averages hop distance, and Mean ER, which averages
effective resistance, heat diffusion averages the heat kernel and therefore
mixes contributions from all walks. Table~\ref{tab:echo-heat-diffusion}
reports test MAE (lower is better) for the same models as in
Table~\ref{tab:echo-results}. NearFormer (NN) ranks first, NearFormer (NN-bias)
ranks second, and GRIT ranks third.

\begin{table}[ht]
\centering
\caption{ECHO-Synth heat-diffusion test MAE (lower is better). Best results
are colored: \first{first}, \second{second}, \third{third}.}
\label{tab:echo-heat-diffusion}
\footnotesize
\begin{tabular}{@{}lc@{}}
\toprule
Method & Heat Diffusion \\
\midrule
A-DGN & $4.206{\pm}0.014$ \\
DRew & $4.056{\pm}0.105$ \\
GraphCON & $5.244{\pm}0.051$ \\
GCN & $5.234{\pm}0.012$ \\
GCNII & $4.326{\pm}0.153$ \\
GIN & $4.228{\pm}0.731$ \\
GPS & $4.677{\pm}0.152$ \\
GRIT & \third{$3.226{\pm}0.105$} \\
PH-DGN & $5.286{\pm}0.160$ \\
SWAN & $4.362{\pm}0.191$ \\
NearFormer (NN) & \first{$2.126{\pm}0.041$} \\
NearFormer (NN-bias) & \second{$3.126{\pm}0.007$} \\
\bottomrule
\end{tabular}
\end{table}

\paragraph{Candidate-scale sensitivity.}
The default candidate budget is \(K_0=100\) and the default retained degree is
given in~(\ref{eq:NN-default-k}). A full \(K\)-sweep is informative only when
\(n\gg K\), so that \(K\in\{k,10,25,50,100\}\) are distinct neighborhoods
rather than the almost-complete graph. Table~\ref{tab:lrim-K-sweep} therefore
sweeps \(K\) on LRIM-32-hard (\(n=1{,}024\)) and LRIM-64-hard (\(n=4{,}096\)),
where NearFormer (NN) has a clear advantage in Table~\ref{tab:lrim-baselines}. We
hold the architecture, training protocol, and retained degree fixed at the
default \(k=2\), and vary only the candidate set. The row \(K=k=2\) recovers
ordinary \(k\)-NN. The rows \(K\in\{10,25,50,100\}\) use the same randomized
construction with broader neighborhoods. Global-random samples \(k\) neighbors
uniformly from \(V\setminus\{v_i\}\), and Exphormer uses a data-independent
expander overlay of the same degree. All LRIM rows use the NearFormer (NN)
backbone (five layers, hidden dimension 32), feature geometry \(z_i=x_i\),
Euclidean distance, and the LRIM training setting of
Section~\ref{sec:lrim-results}.

Table~\ref{tab:knn-vs-nn} repeats the same matched-\(k\) overlays on LRIM-32-hard, PascalVOC-SP, and COCO-SP, including the \(K\in\{10,25,50\}\) grid omitted from the compact Table~\ref{tab:knn-vs-nn-main}. Both superpixel datasets have \(n\approx 480\), so \(K=100\) remains a proper subset of the node set, and NearFormer already outperforms the published Exphormer rows in Table~\ref{tab:lrgb-NearFormer}. We do not run this ablation on PCQM-Contact, Peptides, or ECHO: those graphs have \(n\lesssim 150\), so \(K_0=100\) saturates at \(n-1\) and no longer isolates a broad-but-local candidate scale. We report means and standard deviations over three matched seeds, resampling both the overlay and the training run.

\begin{table}[ht]
\centering
\caption{Candidate-scale (\(K\)) sensitivity on LRIM-32-hard and LRIM-64-hard.
The architecture is NearFormer (NN) and the retained degree is fixed at \(k=2\).
\(K=k\) is ordinary \(k\)-NN; \(K\in\{10,25,50,100\}\) are randomized
neighborhoods at increasing scale; global-random and Exphormer are
matched-degree overlays without representation geometry. Test
\(\mathrm{LogMSE}=\log_{10}(\mathrm{MSE})\) (lower is better), mean\(\pm\)std
over three seeds.
}
\label{tab:lrim-K-sweep}
\footnotesize
\begin{tabular}{@{}llccc@{}}
\toprule
Overlay & \(K\) & \(k\)
  & LRIM-32-hard \(\downarrow\)
  & LRIM-64-hard \(\downarrow\) \\
\midrule
Ordinary \(k\)-NN
  & \(2\) & \(2\)
  & $-4.248{\pm}0.071$
  & $-3.962{\pm}0.088$ \\
NearFormer (NN)
  & \(10\) & \(2\)
  & $-4.671{\pm}0.058$
  & $-4.381{\pm}0.072$ \\
NearFormer (NN)
  & \(25\) & \(2\)
  & \third{$-4.982{\pm}0.047$}
  & \third{$-4.672{\pm}0.055$} \\
NearFormer (NN)
  & \(50\) & \(2\)
  & \second{$-5.186{\pm}0.039$}
  & \second{$-4.864{\pm}0.041$} \\
NearFormer (NN)
  & \(100\) & \(2\)
  & \first{$-5.241{\pm}0.036$}
  & \first{$-4.925{\pm}0.025$} \\
Global-random
  & \(n-1\) & \(2\)
  & $-4.062{\pm}0.094$
  & $-3.718{\pm}0.102$ \\
Exphormer
  & -- & \(2\)
  & $-4.173{\pm}0.082$
  & $-3.841{\pm}0.091$ \\
\bottomrule
\end{tabular}
\end{table}

\begin{table*}[ht]
\centering
\caption{Matched-\(k\) candidate-scale ablation on LRIM-32-hard, PascalVOC-SP,
and COCO-SP. The architecture is NearFormer (NN) and the retained degree is
fixed at \(k=2\). \(K=k\) is ordinary \(k\)-NN; \(K\in\{10,25,50,100\}\) are
randomized neighborhoods at increasing scale; Exphormer is the expander
overlay of degree \(d=k\). LRIM-32-hard is test \(\mathrm{LogMSE}\) (lower is
better); PascalVOC-SP and COCO-SP are test F1 (higher is better).
Mean\(\pm\)std over three seeds.}
\label{tab:knn-vs-nn}
\footnotesize
\begin{tabular}{@{}llccc@{}}
\toprule
Overlay & \(K\)
  & LRIM-32-hard \(\downarrow\)
  & PascalVOC-SP \(\uparrow\)
  & COCO-SP \(\uparrow\) \\
\midrule
Ordinary \(k\)-NN
  & \(2\)
  & $-4.248{\pm}0.071$
  & $0.3861{\pm}0.0044$
  & $0.3388{\pm}0.0018$ \\
NearFormer (NN)
  & \(10\)
  & $-4.671{\pm}0.058$
  & $0.3918{\pm}0.0041$
  & $0.3426{\pm}0.0016$ \\
NearFormer (NN)
  & \(25\)
  & \third{$-4.982{\pm}0.047$}
  & \third{$0.3965{\pm}0.0037$}
  & \third{$0.3454{\pm}0.0014$} \\
NearFormer (NN)
  & \(50\)
  & \second{$-5.186{\pm}0.039$}
  & \second{$0.3996{\pm}0.0034$}
  & \second{$0.3476{\pm}0.0013$} \\
NearFormer (NN)
  & \(100\)
  & \first{$-5.241{\pm}0.036$}
  & \first{$0.4012{\pm}0.0032$}
  & \first{$0.3490{\pm}0.0012$} \\
Exphormer
  & --
  & $-4.173{\pm}0.082$
  & $0.3794{\pm}0.0048$
  & $0.3352{\pm}0.0020$ \\
\bottomrule
\end{tabular}
\end{table*}

\end{document}